\documentclass[sigconf, ann]{acmart}
\usepackage{subfig}
\usepackage{amsthm}
\usepackage{hyperref}
\usepackage{algorithm}
\usepackage{algorithmic}
\usepackage{bm}
\usepackage{color}
\usepackage{colortbl}
\usepackage{booktabs}
\usepackage{multirow} 

\AtBeginDocument{%
	\providecommand\BibTeX{{%
			\normalfont B\kern-0.5em{\scshape i\kern-0.25em b}\kern-0.8em\TeX}}}
\copyrightyear{2023}
\acmYear{2023}
\setcopyright{acmlicensed}
\acmConference[MM '23] {Proceedings of the 31st ACM International Conference on Multimedia}{October 29--November 3, 2023}{Ottawa, ON, Canada.}
\acmBooktitle{Proceedings of the 31st ACM International Conference on Multimedia (MM '23), October 29--November 3, 2023, Ottawa, ON, Canada}
\acmPrice{15.00}
\acmISBN{979-8-4007-0108-5/23/10}

\acmDOI{10.1145/3581783.3611727}

\begin{document}

\title{WaterFlow: Heuristic Normalizing Flow for Underwater Image Enhancement and Beyond}
\author{Zengxi Zhang}
\affiliation{%
	\institution{Dalian University of Technology}
	\country{}
}
\email{cyouzoukyuu@gmail.com}

\author{Zhiying Jiang}
\affiliation{%
	\institution{Dalian University of Technology}
	\country{}
}
\email{zyjiang0630@gmail.com}

\author{Jinyuan Liu}
\affiliation{%
	\institution{Dalian University of Technology}
	\country{}
}
\email{atlantis918@hotmail.com}

\author{Xin Fan}
\affiliation{%
	\institution{Dalian University of Technology}
	\country{}
}
\email{xin.fan@dlut.edu.cn}

\author{Risheng Liu}
\affiliation{%
	\institution{Dalian University of Technology \country{}}
	\institution{Peng Cheng Laboratory \country{}}
}
\email{rsliu@dlut.edu.cn}
\authornote{Corresponding author: Risheng Liu.}
\begin{abstract}
  Underwater images suffer from light refraction and absorption, which impairs visibility and interferes the subsequent applications. Existing underwater image enhancement methods mainly focus on image quality improvement, ignoring the effect on practice. To balance the visual quality and application, we propose a heuristic normalizing flow for detection-driven underwater image enhancement, dubbed WaterFlow. Specifically, we first develop an invertible mapping to achieve the translation between the degraded image and its clear counterpart. Considering the differentiability and interpretability, we incorporate the heuristic prior into the data-driven mapping procedure, where the ambient light and medium transmission coefficient benefit credible generation. Furthermore, we introduce a detection perception module to transmit the implicit semantic guidance into the enhancement procedure, where the enhanced images hold more detection-favorable features and are able to promote the detection performance. Extensive experiments prove the superiority of our WaterFlow, against state-of-the-art methods quantitatively and qualitatively.
\end{abstract}



\maketitle

\section*{}
\begin{figure}[!htb]
	\centering
	\setlength{\tabcolsep}{1pt}
	\vspace{-18pt}
	\includegraphics[width=0.5\textwidth]{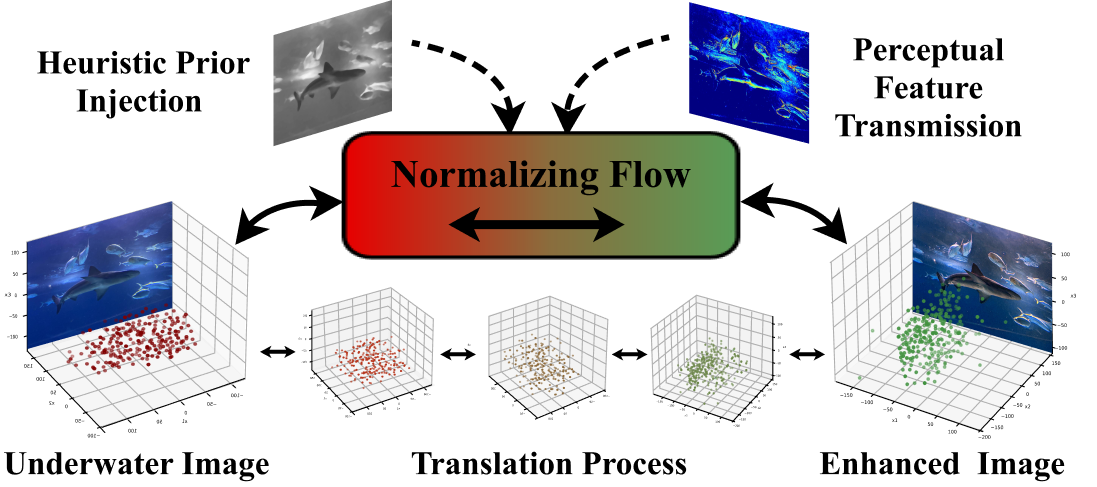}
	\vspace{-18pt}
	\caption{The invertible translation process between underwater and
			enhanced images. Heuristic priors are embedded to enhance the interpretability of the enhanced network. High-level perceptual information is introduced into the model by backpropagation to help the enhanced image potentially contain more semantic information conducive to object detection. We illustrate the k-means clustering results along with the procedure of the invertible translation.}
	\vspace{-13pt}
	\label{fig:firstfigure}
\end{figure}
\section{Introduction}

In recent years, there have been significant advances in underwater robots for exploration in various fields~\cite{lin2020underwatersearch,yoerger2021underwaterinvestigation}. However, underwater object detection, a critical component of underwater exploration tasks, still faces considerable difficulties. The images captured underwater often suffer from severe distortion due to the complex and variable underwater environment, leading to a significant reduction in image visibility. The performance of subsequent object detection applications is also significantly affected.

To mitigate the impact of image distortion on underwater object detection, existing methods typically employ underwater image enhancement as a preprocessing step for object detection~\cite{chen2020enhancedetection}.
They feed the enhanced underwater images as prospective guidance into the detection module to achieve better detection accuracy. Traditional underwater enhancement methods estimate imaging parameters through the underwater degradation formula to obtain enhanced images. However, fixed imaging parameters may not fully simulate the diverse and complex nature of real underwater environments. In recent years, deep learning has been widely concerned by researchers. They~\cite{li2021ucolor,jiang2022topal} have achieved good restoration effects through the manually crafted network by end-to-end training. However, image enhancement and object detection are usually regarded as two parallel independent tasks. Enhancement results that solely focus on visual perception are insufficient to fully capture the scene information required by subsequent detection algorithms. Moreover, the enhancement network may introduce uncertain interferences that affect the detection performance. 

In this paper, we propose a detection-driven heuristic normalizing flow for underwater image enhancement. 
we first develop a reversible translation framework based on normalizing flow to facilitate domain translation between degraded underwater images and clear counterpart. 
Specifically, the reversible translation between them is established by synchronous optimization with shared parameters and bilateral constraints of forward and reverse processes. The forward process aims to map the degraded image to the clear restored image by learning the nonlinear function $F_n(.)$, and the reverse process aims to map the clear image back to the degraded image through the nonlinear function $F^{-1}_n(.)$. 
Then, considering the differentiability and interpretability, we estimate the underwater imaging parameters using different modalities of knowledge and incorporate them as heuristic priors into the data-driven mapping process. Guided by the underwater imaging physical model, the proposed method effectively prevents the introduction of the noise interference and undesirable artifacts, thereby avoiding adverse impacts on subsequent detection tasks.
In order to improve the adaptability of the enhanced results for subsequent detection tasks, we further introduce a Detection Perception Module. By propagating the high-level perceptual features to the enhancement module, the generated enhanced images can implicitly possess more semantic information beneficial for subsequent detection tasks while achieving visually pleasing enhancement effects.

In summary, contributions can be concluded as follows:
\begin{itemize}
	\item We apply the normalizing flow to the underwater image enhancement task, which realizes the invertible mapping between the degradation image and its clear counterpart.
	\item We incorporate the heuristic prior into the data-driven mapping process, which can be widely applied in a variety of real underwater scenes by improving the interpretability of the whole enhancement framework.
	\item We propose a Detection Perception Module, which transmits high-level latent perceptual information to retain and extract detection-oriented semantic information.
	\item Qualitative and quantitative results demonstrate that the proposed WaterFlow recovers the intrinsic scene clearly and is more conducive to the subsequent detection.
\end{itemize}

\section{RELATED WORK}
\subsection{Underwater Image Enhancement}
In recent years, numerous underwater image enhancement methods have been proposed. The early traditional methods often adjust the pixel distribution of different color channels to weaken the degradation of natural light underwater. 
Ghani~\emph{et al.}~\cite{ghani2015underwater} extended the histogram of the red channel and the blue channel of the underwater image upward and downward respectively according to the law of Rayleigh distribution. Li~\emph{et al.}~~\cite{li2016single} corrected the red channel according to the Gray-World assumption theory after the color restoration of the blue-green image through the dark channel. Although these methods can weaken the fading of light on different color channels, it is easier to introduce underwater images into artificial colors due to the lack of guidance from physical models. 
Therefore, model-based methods~\cite{li2017hybrid,wang2017single,peng2018generalization,yang2023implicit} are also widely used to improve the interpretability of networks by incorporating domain-specific prior knowledge.

In recent years, many methods based on deep learning have been proposed: 
Li~\emph{et al.}~\cite{li2021ucolor} introduced the multi-color space into the transmission-guided network to reduce the influence of color casts on underwater images. 
Mu~\emph{et al.}~\cite{mu2022structure} introduced a bi-level model that hierarchically incorporates various knowledge modalities to enhance the quality of underwater images.
Jiang~\emph{et al.}~\cite{jiang2022topal} proposed a perceptual adversarial mechanism and introduced the global module to narrow the gap between enhanced images and reference images. 
Huang~\emph{et al.}~\cite{huang2023contrastive} proposed a semi-supervised framework based on the mean-teacher approach to enhance the generalization ability of the enhance model on real-world data.
\subsection{Normalizing Flow}
Normalizing flow facilitates the transformation between complicated probability distribution and Gaussian distribution through bijection functions and differentiable mappings. The normalizing flow has received far less attention than GANs and VAEs because it requires a significant quantity of video memory with a fine structural design to handle a large amount of reversible computing. Many methods for accelerating the calculation have been proposed in recent years. Dinh~\emph{et al.}~\cite{dinh2014nice} proposed the additive coupling layer, which makes the flow easier to calculate the determinant of Jacobian matrix. In order to increase the log-likelihood, Kingma~\emph{et al.}~\cite{kingma2018glow} developed the actnorm to normalize each channel of input features and $1 \times 1$ convolution in place of the permutation~\cite{dinh2014nice}. Conditional normalizing flow was introduced for increasing the expressiveness and flexibility of normalizing flows, and has been widely used in vision tasks~\cite{lugmayr2020srflow,an2021artflow,abdal2021styleflow,li2021dehazeflow,wang2022llflow,jiang2023modality}.
However, there is little precedent dedicated to adopting flow-based methods to solve the problem of ill-posed underwater image enhancement tasks.

\subsection{Object Detection} 
In recent years, due to the rapid development of deep learning, the effect of object detection has also been significantly improved. The existing methods are generally divided into one-stage method and two-stage method. The two-stage method~\cite{girshick2014rich,ren2015faster,he2017mask,sun2021sparse,cheng2022implicit} first needs to extract the region of interest proposals area, and then classify it through the classification of neural network. By contrast, the classification label and boundary information are often directly obtained by the one-stage method~\cite{redmon2018yolov3,kim2020paa,feng2021tood,xie2022pyramid}. 

The current detectors can achieve good performance when facing in-air images. However, the problems of scene blur and light imbalance faced by underwater images will greatly affect the effectiveness of the detectors. Existing studies~\cite{liu2021searching,liu2022target,Ma2022LL,liu2023bi,liu2023task} have used low-level visual enhancement modules as preprocessing steps for detection tasks. However, the existing enhanced networks for the underwater image often do not play a significant role in the detector, which ignores the position and semantic information while restoring the contrast of the underwater image.

\begin{figure*}[tp]
	\centering
	\setlength{\tabcolsep}{1pt}
	\includegraphics[width=\textwidth]{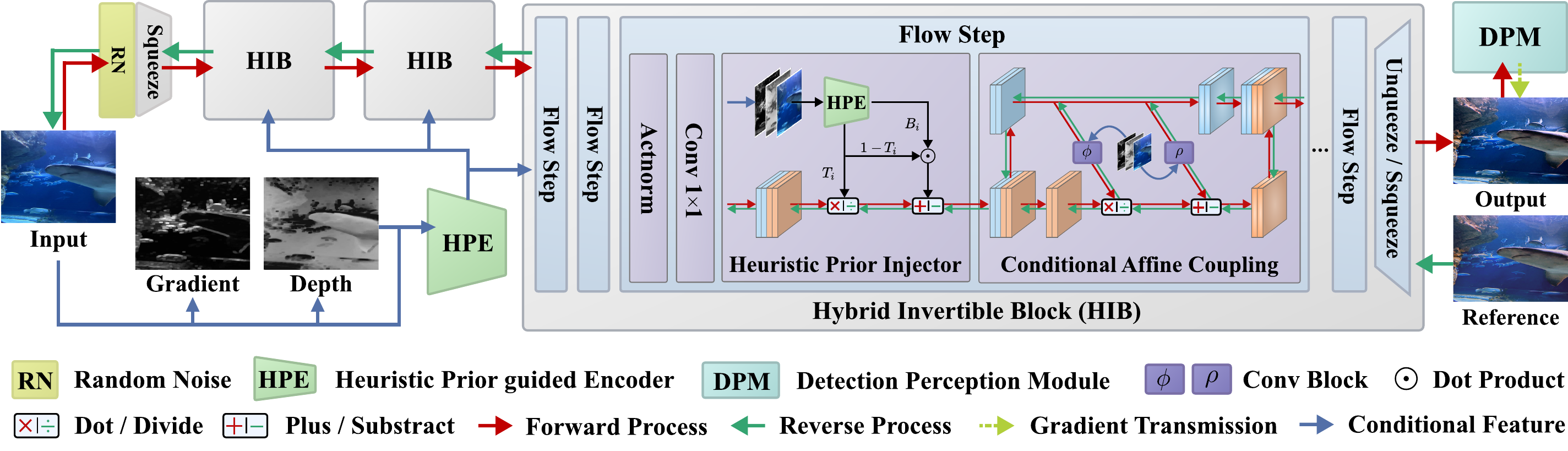}
	\vspace{-20pt}
	\caption{Workflow of the proposed WaterFlow. The underwater imaging physical model is embedded into the normalizing flow as a heuristic prior to better simulate the mapping of the underwater image and its clear counterpart. At the same time, the detection module is combined to improve the detection effect of underwater object detection by transmitting the high-level latent feature to the enhancement module.}
	\vspace{-5pt}
	\label{fig:workflow}
\end{figure*}

\section{The Proposed Method}
Existing deep learning based underwater image enhancement methods rely on the enriched training data for a more robust and effective performance. However, the ignorance of physical priors in the learning process introduces the noise interference and undesirable artifacts, degrading the visibility further and undermining the subsequent detection performance. To overcome the above limitations, we propose a detection-driven heuristic normalizing flow network for underwater image enhancement. Specifically, to reduce the dependence on training data, we develop a normalizing flow based reversible translation framework to achieve domain translation between degraded underwater images and clear counterpart. The heuristic model constraints incorporated with gradient and depth information are embedded into the reversible procedure, guiding the reversible results to be more accurate and reliable. Moreover, to improve the applicability of the enhanced results for subsequent object detection task, we propose a Detection Perception Module that feeds back the performance of the enhanced results in detection tasks, allowing the enhancement module to learn more detection-favorable implicit features. Next, we provide the detailed description of each module.

\subsection{Hybrid Invertible Block}
Hybrid Invertible Block (HIB) was introduced as the main part of the heuristic normalizing flow. It ensembles the heuristic prior into the data-driven network to build the invertible mapping relationship between the underwater image and its clear counterpart. As illustrated in Fig.~\ref{fig:workflow}, the underwater image restoration and the underwater image degradation process are combined in a invertible mapping manner during the training process. In the forward process, the underwater image is squeezed and input into multiple invertible blocks to generate enhanced images. In the reverse process, clear images are inversely input to invertible blocks with shared parameters to generate degraded images. 

The Hybrid Invertible Block consists of the following parts: 
Actnorm~\cite{kingma2018glow}, Conv 1 $\times$1~\cite{kingma2018glow}, Heuristic Prior Injector, Conditional Affine Coupling~\cite{li2021dehazeflow} and squeeze operation.

Actnorm~\cite{kingma2018glow} is a normalizing operation, which can change the input tensor into the zero mean and unit variance tensors. 
Conv $1 \times 1$~\cite{kingma2018glow} can convolve the input tensor of $c \times h \times w$ by the weight matrix of $c \times c$ to make the calculation of Jacobian matrix and network inversion easier.  

Heuristic Prior Injector was designed to better characterize the underwater imaging model by embedding a heuristic prior, which will be introduced in detail in section~\ref{HPI}.

Conditional Affine Coupling was proposed by~\cite{li2021dehazeflow}, which realizes the mutual conversion with conditional information between output and input through reversible transformation. We modified the previous formulation for adjusting the underwater image enhancement framework, which is described as:

\begin{equation}
	\begin{aligned}
		\mathbf{u}^1_{i+1}=&~\mathbf{u}^1_i, \\
		\mathbf{u}^2_{i+1}=& \left(\phi_i\left(\mathcal{C}(\mathbf{u}^1_i, f^{A}_i\left(I_u,I_g,I_d\right))\right)\right) \odot \mathbf{u}^2_i\\
		&+\rho_i\left(\mathcal{C}(\mathbf{u}^1_i, f^{B}_i\left(I_u,I_g,I_d\right))\right),
	\end{aligned}
\end{equation}
where $\mathbf{u}_{i+1}= \mathcal{C}(\mathbf{u}_{i+1}^1,\mathbf{u}_{i+1}^2)$,~$\mathcal{C}$ represents the concatenation operation. $\phi_i$,~$\rho_i$,~$f^{A}_i$,~ and $f^{B}_i$ denote convolutional networks.~$i \in \{1,...,n-1\}$ represents the $i$-th flow steps,~$I_u$, $I_g$ and $I_d$ will be explained in detail in section~\ref{HPI}.
Squeeze operation can convert the image of $c \times h \times w$ to $4 c \times \frac{h}{2} \times \frac{w}{2}$. Unsqueeze is the reverse operation of squeeze.
By the combination of these operations, the proposed architecture can attain accurate mapping between the underwater image and it clear counterpart while maintaining reversibility.

\subsection{Heuristic Prior Injector}\label{HPI} 
The Heuristic Prior guided Injector (HPI) is designed to incorporate the estimated physical imaging parameters into the invertible block. According to the underwater image formation model~\cite{drews2013formationmodel,chiang2011underwater}, the enhanced image can be represented as:
\begin{equation}
	J^c(x) = \frac{1}{t(x)}I^c(x) + \frac{1}{t(x)}B^c(t(x)-1), c \in\{r, g, b\},
	\label{eq:imaginginverse}
\end{equation}
where $I$ and $J$ denote the underwater image captured by sensors and the enhanced image at pixel point $x$ respectively.~$c \in\{r, g, b\}$ represents the corresponding color channels.~$B$ and $t$ indicates the ambient light and the medium transmission coefficient.
$t$ can also be described as $t(x)=e^{\beta d(x)}$ with the scene depth $d(x)$ and the attenuation coefficient $\beta$ according to the Beer-Lambert law~\cite{bouguer1729essai}.

Considering that the gradient map provides information about the edges and contrast in the image, which can be used to estimate the scatteration of light by the atmosphere and fine impurities in the water. Furthermore, the depth map provides information about the attenuation of the light propagation, which is also a key factor in the Beer-Lambert law. Inspired by~\cite{peng2018generalization}, we first estimate the depth map $I_d$ and the gradient map $I_g$ as auxiliary information from the underwater image.

Then, we concatenate them with underwater image as input to the Heuristic Prior guided Encoder (HPE) $\mathcal{H}$ to progressively estimate the imaging parameters by extracting the depth, gradient and color information of the underwater image. The equation can be illustrated as:
\begin{equation}
	B_i,T_i =\mathcal{S}\left(\mathcal{H}_i\left(\mathcal{C}\left(I_d,I_g,I_u\right)\right)\right),
\end{equation}
where $I_u$ represents underwater images.~$\mathcal{C}$ and $\mathcal{S}$ denote the concatenate and split operation respectively.~$T$ is the reciprocal of $t$. The detailed architecture of $\mathcal{H}$ is in the appendix due to limited space. After estimating $T$ and $B$, we insert them into the HIB as heuristic information.
The operation can be formulated as:
\begin{equation}
	\begin{aligned}
		\mathbf{u}_{i+1} =&~T_i \odot \mathbf{u}_i + B_i\odot(1-T_i),\\
		\mathbf{u}_{i} =&\left(\mathbf{u}_{i+1} - B_i\odot(1-T_i)\right) / T_i,\\
	\end{aligned}
\end{equation}
where $\mathbf{u}_i$ represents the middle feature of the $i$-th flow steps, $\odot$ denotes dot product. By introducing the underwater image formation model based heuristic prior to the invertible blocks, the proposed method can generate enhanced images, which are more suitable for realistic underwater scenes.

\subsection{Detection Perception Module}
Object detection aims to acquire the location and category of each object. Hence, the performance of object detection will be improved if the enhanced image holds a more profound level of semantic feature and object localization information.
Inspired by the perceptual loss~\cite{johnson2016perceptual}, deep neural networks designed for high-level visual tasks can retain semantic information and own the ability to describe potential features. Therefore, in order to make the enhanced images potentially improve the effectiveness of the underwater object detection, we propose a Detection Perception Module (DPM) to improve the performance of the enhancement module in preserving and extracting detection-oriented perceptual features.

In the whole training process, we first use the enhancement module as the data preprocessing stage of underwater object detection. After that, the enhanced image was directly fed into the DPM and conducted separate detection training. Then, we jointly optimize the pre-trained DPM and enhancement module through different benchmarks. Specifically, the underwater image is first input to the enhancement module to obtain enhanced data. We input it into the detection module to transmit the high-level implicit perceptual features extracted from the network to the enhancement module, so as to guide the visual improvement effect to be more conducive to subsequent object detection tasks.

\subsection{Loss Function}
Contrastive learning has been applied to multiple low-level visual tasks~\cite{liu2022tacl,liu2022coconet,liu2023holoco,jiang2023contrastive}. We introduce contrastive learning to improve the quality of enhanced images by making them closer to the in-air images and more isolated from the underwater images. We have designated the reference image as a positive sample and the original underwater image as a negative sample. VGG19~\cite{simonyan2014vgg19} was used to extract the implicit characteristic of enhanced images.  The contrastive loss is expressed as follows:

\begin{equation}
	\mathcal{L}_{c}=\sum_{i=1}^N \rho_i 
	\cdot
	\frac{\left\|\mathcal{V}_i(I_r)-\mathcal{V}_i\left(G_E(I_u)\right)\right\|_1}
	{\left\|\mathcal{V}_i(I_u)-\mathcal{V}_i\left(G_E(I_u)\right)\right\|_1},
\end{equation}
where $\mathcal{V}_i$ represents the $i$-th layer of VGG19~\cite{simonyan2014vgg19}, and $\rho_i$ denotes the $i$-th weight of each layer. $G_E$ represents the proposed enhancement network. $I_u$ and  $I_r$ represent the underwater image and the reference image respectively.

We additionally incorporate the style loss~\cite{deng2022stytr2} to make the generated output closer to the style pattern of the reference image. The definition of style loss function $\mathcal{L}_s$ is described as follows:
\begin{equation}
	\begin{aligned}
		\mathcal{L}_s= & \frac{1}{N} \sum_{i=0}^{N}\left\|\mu\left(\mathcal{V}_i\left(G_E(I_u)\right)\right)-\mu\left(\mathcal{V}_i\left(I_r\right)\right)\right\|_2 \\
		& +\left\|\nu\left(\mathcal{V}_i\left(G_E(I_u)\right)\right)-\nu\left(\mathcal{V}_i\left(I_r\right)\right)\right\|_2,
	\end{aligned}
\end{equation}
where $\mu$ and $\nu$ represent the mean and variance of the image.~$N$ is the number of layers.

Localization loss $\mathcal{L}_{\text{loc}}$ and classification loss $\mathcal{L}_{\text{cla}}$ are used as detection-driven loss $\mathcal{L}_{\text {det}}$, which is illustrated as:
\begin{equation}
	\mathcal{L}_{\text {det}}=\mathcal{L}_{\text {cla }}+\mathcal{L}_{\text {loc }},
\end{equation}
where classification loss is designed to minimize the discrepancy between the prediction category and the ground truth category. The localization loss is adopted to reduce the position difference between the prediction box and the ground truth box. We employ Focal loss~\cite{lin2017focalloss} and GIoU loss~\cite{rezatofighi2019giou} as classification loss and localization loss respectively.

Ideally, the incorporation of the reverse mapping process could impose a regular constraint on the degraded image, thereby enhancing the performance of the forward mapping~\cite{guo2020closed}. Therefore, L1 loss is used as a bilateral constraint to make the output of the forward and the reverse process closer to the reference and underwater image respectively. The formula can be described as:
\begin{equation}
	\mathcal{L}_{1}=\left\|G_E(I_u) - I_r\right\|_1 + \left\|G_E^{-1}(I_r) - I_u\right\|_1,
\end{equation}	
where $G_E^{-1}$ denotes the reverse transformation of $G_E$. Therefore, the total loss of network training is expressed as follows:		
\begin{equation}
	\mathcal{L}_{total}=\lambda_1\mathcal{L}_{c}+\lambda_2\mathcal{L}_{s}+\lambda_3\mathcal{L}_{\text {det}}+\lambda_4\mathcal{L}_{1}.
\end{equation}								

\begin{figure*}[!htb]
	\centering
	\setlength{\abovecaptionskip}{0.cm}
	\setlength{\tabcolsep}{1pt}
	\begin{tabular}{cccccccccccc}
		\includegraphics[width=0.122\textwidth]{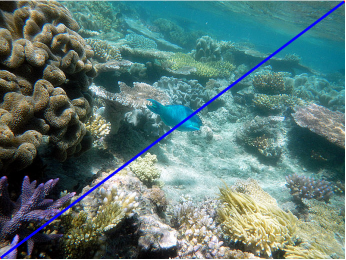}
		&\includegraphics[width=0.122\textwidth]{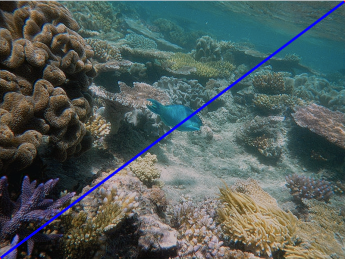}
		&\includegraphics[width=0.122\textwidth]{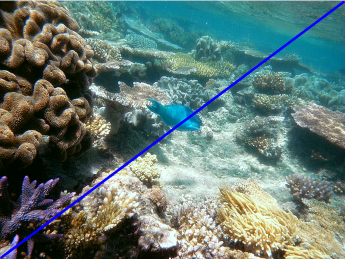}
		&\includegraphics[width=0.122\textwidth]{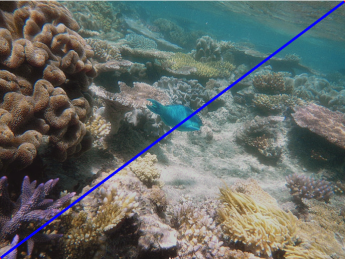}
		&\includegraphics[width=0.122\textwidth]{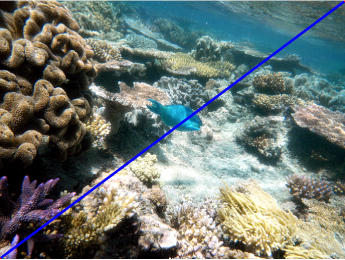}
		&\includegraphics[width=0.122\textwidth]{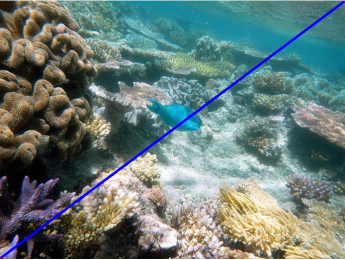}
		&\includegraphics[width=0.122\textwidth]{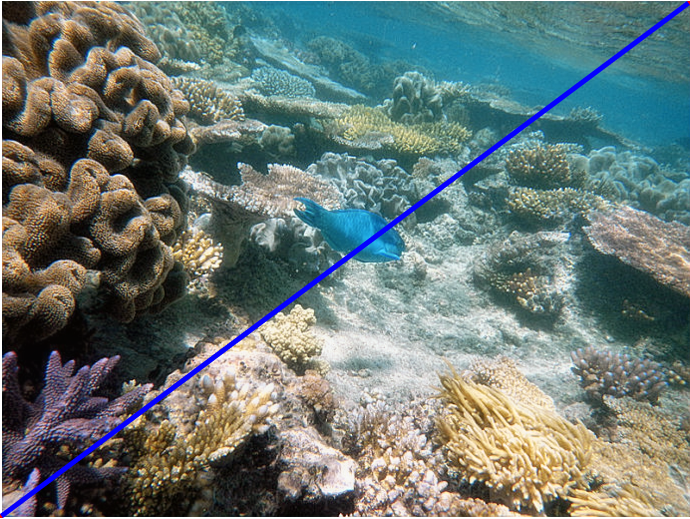}
		&\includegraphics[width=0.122\textwidth]{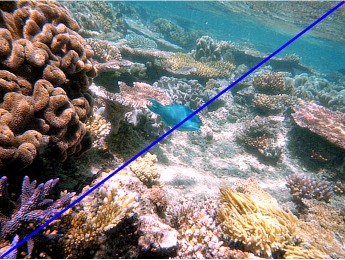}
		\\
		\includegraphics[width=0.122\textwidth,height=0.07\textheight]{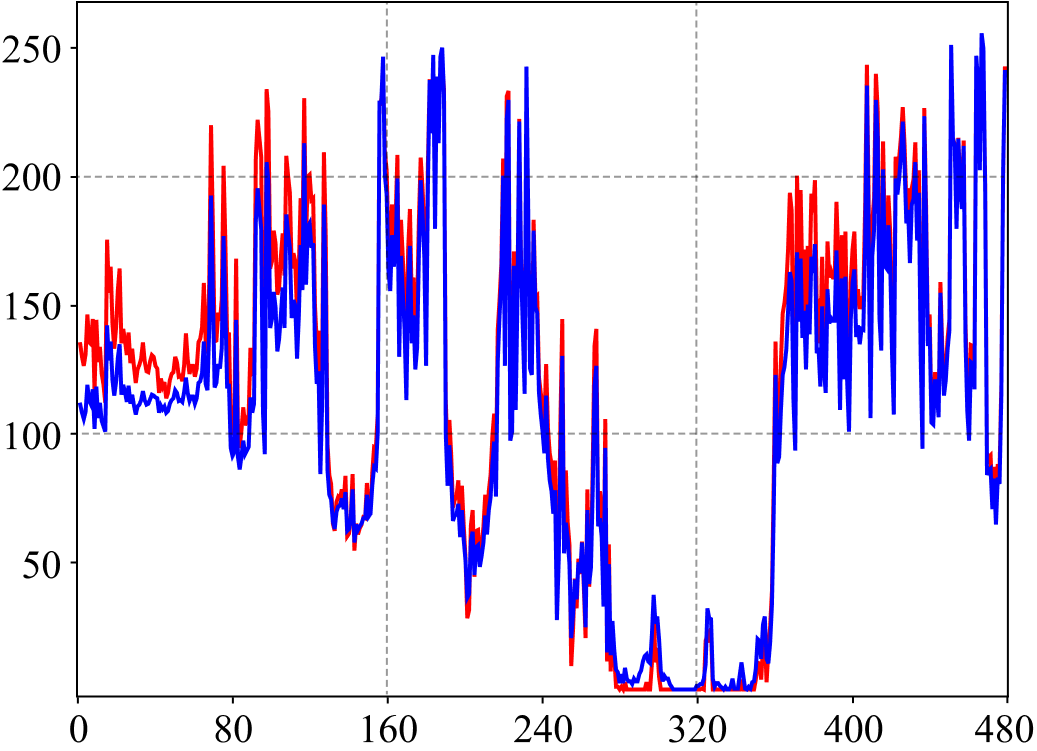}
		&\includegraphics[width=0.122\textwidth,height=0.07\textheight]{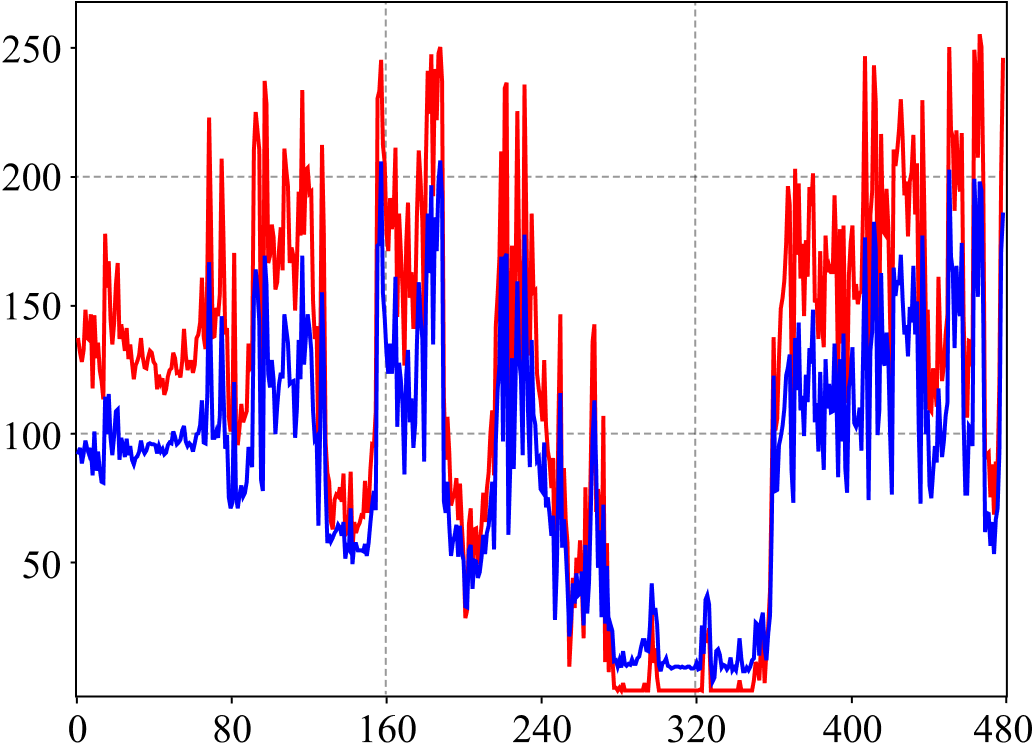}
		&\includegraphics[width=0.122\textwidth,height=0.07\textheight]{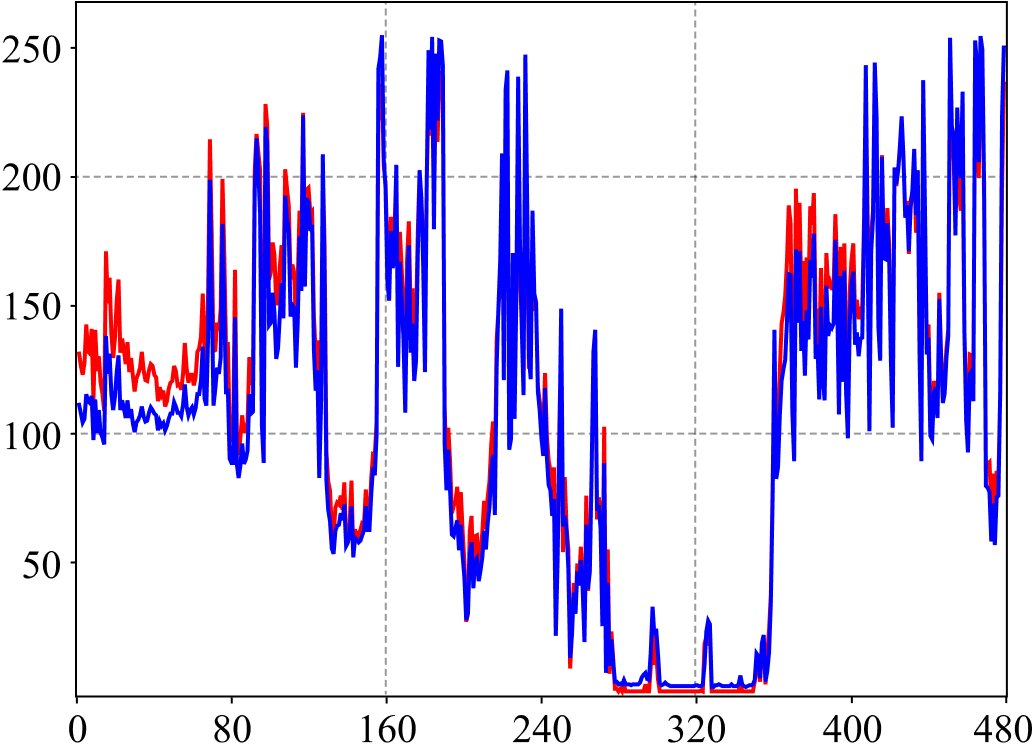}
		&\includegraphics[width=0.122\textwidth,height=0.07\textheight]{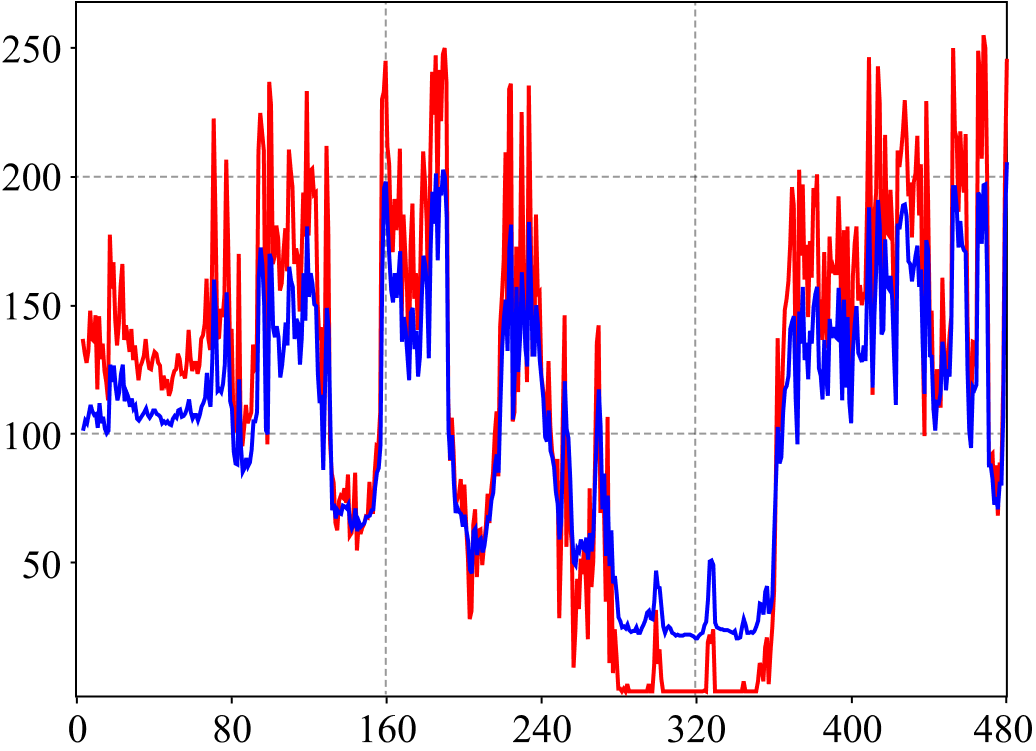}
		&\includegraphics[width=0.122\textwidth,height=0.07\textheight]{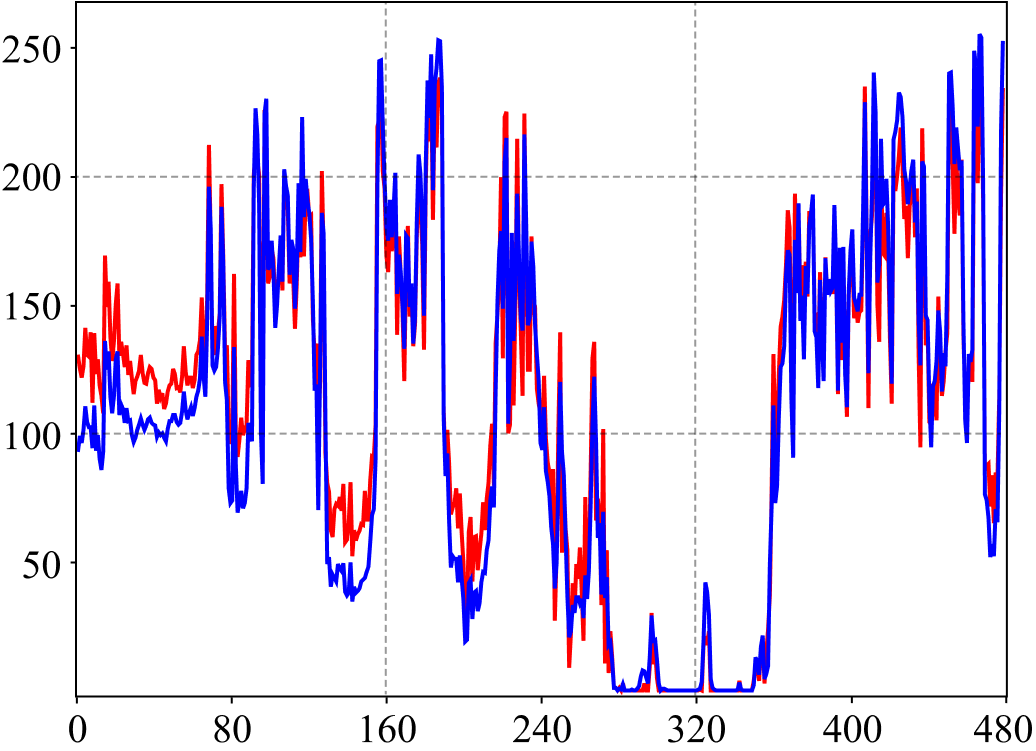}
		&\includegraphics[width=0.122\textwidth,height=0.07\textheight]{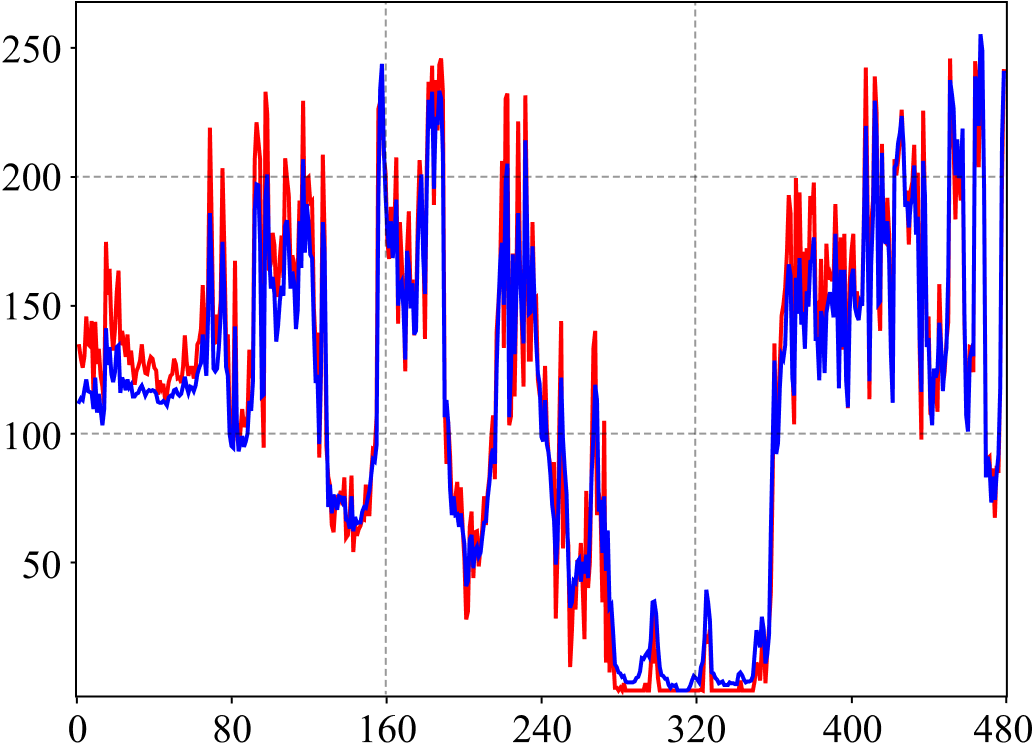}
		&\includegraphics[width=0.122\textwidth,height=0.07\textheight]{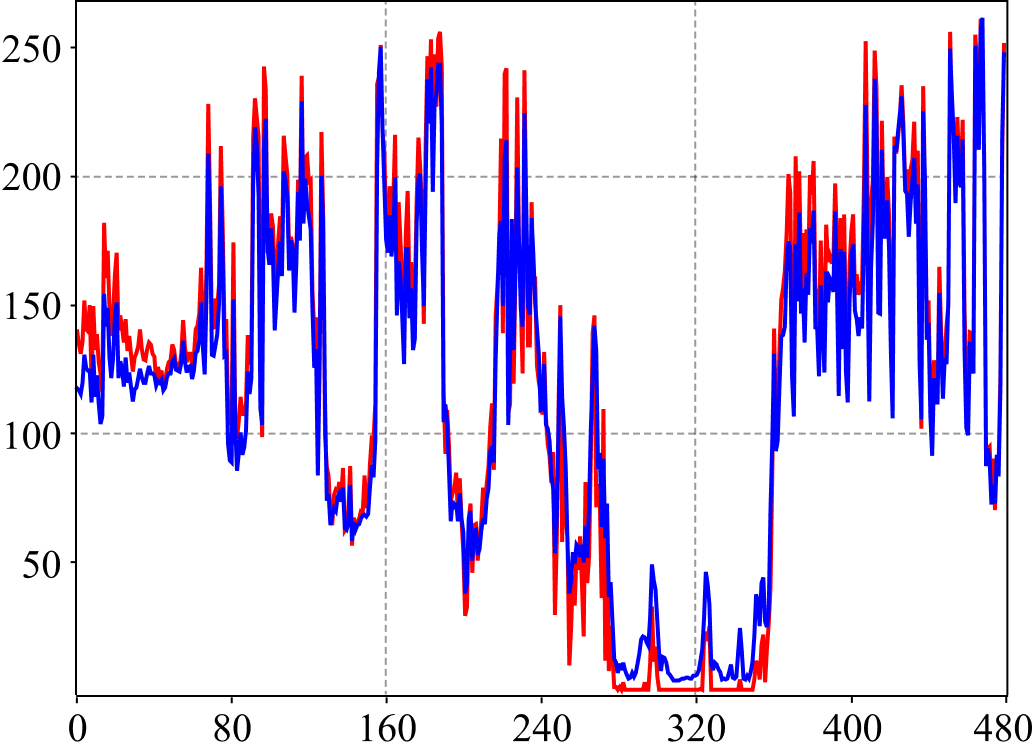}
		&\includegraphics[width=0.122\textwidth,height=0.07\textheight]{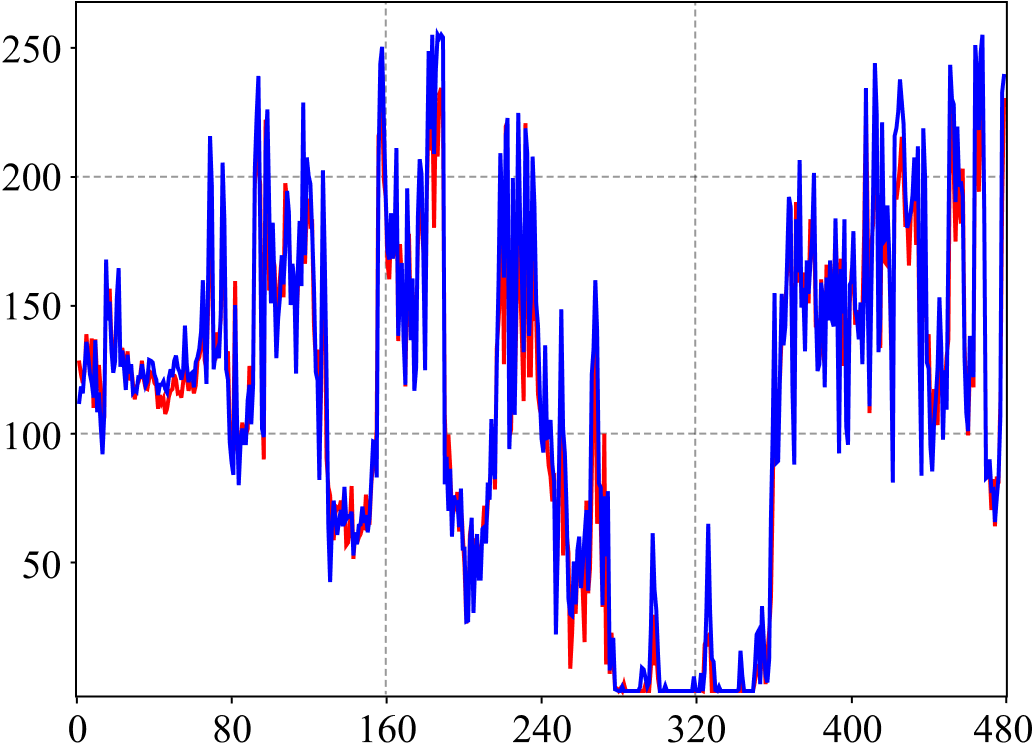}
		\\Input&Water-Net&DLIFM&Ucolor&TOPAL&TACL&SemiUIR&Ours
	\end{tabular}
	\vspace{1pt}
	\caption{Enhancement results on UIEBD dataset. We compared the pixel distribution of the reference images and enhanced images. Obviously, the distribution of our result is the closest to the distribution of reference image.} 
\label{fig:UIEBD}
\end{figure*}
\begin{figure*}[!htb]
\centering
\setlength{\tabcolsep}{1pt}
\begin{tabular}{cccccccccccc}
	\includegraphics[width=0.122\textwidth,height=0.075\textheight]{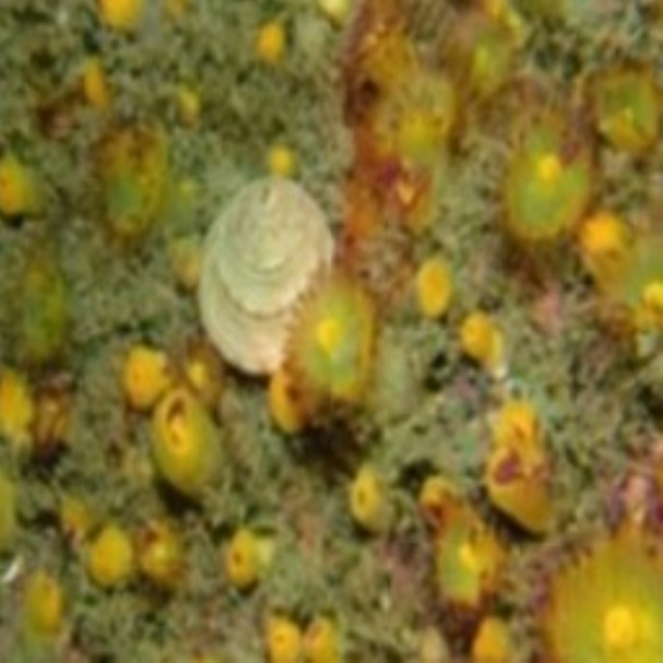}
	&\includegraphics[width=0.122\textwidth,height=0.075\textheight]{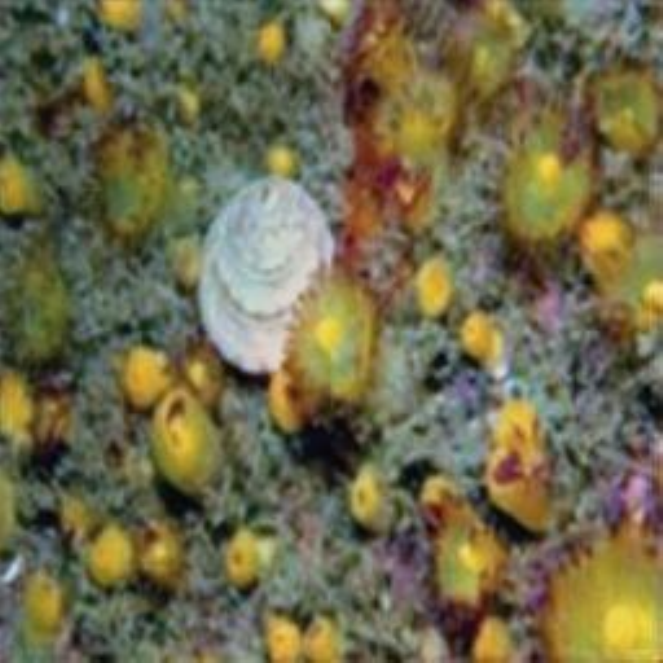}
	&\includegraphics[width=0.122\textwidth,height=0.075\textheight]{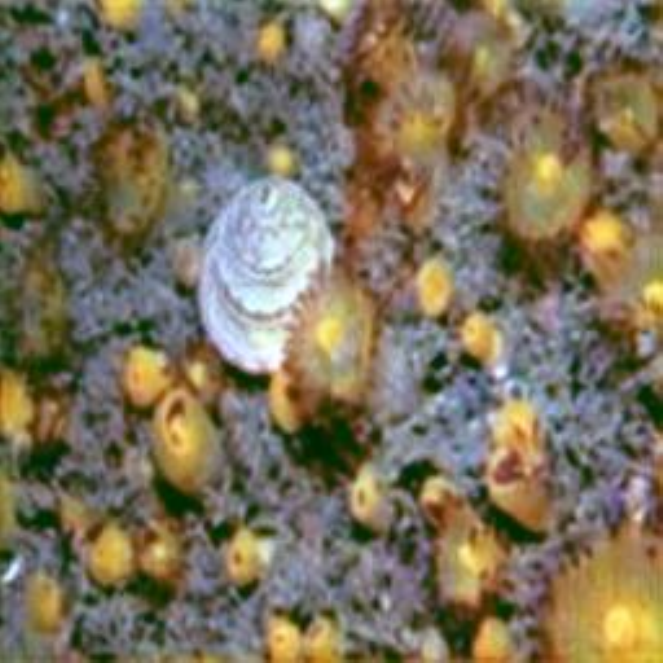}
	&\includegraphics[width=0.122\textwidth,height=0.075\textheight]{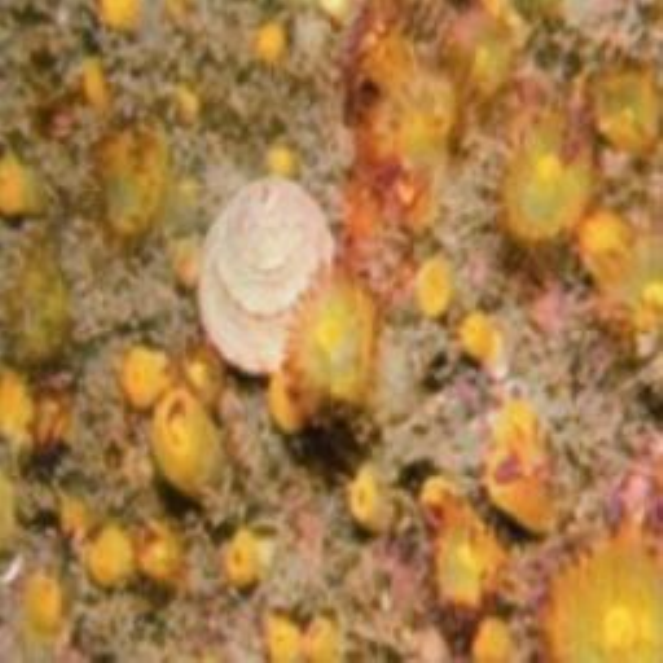}
	&\includegraphics[width=0.122\textwidth,height=0.075\textheight]{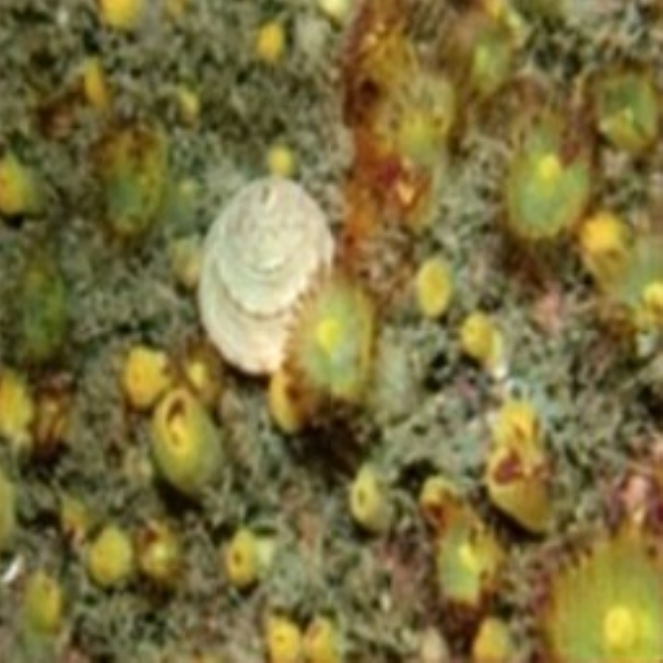}
	&\includegraphics[width=0.122\textwidth,height=0.075\textheight]{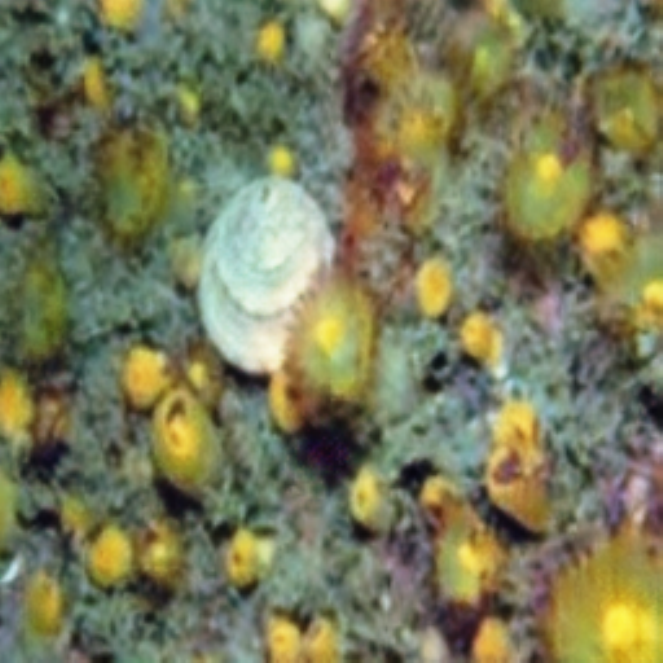}
	&\includegraphics[width=0.122\textwidth,height=0.075\textheight]{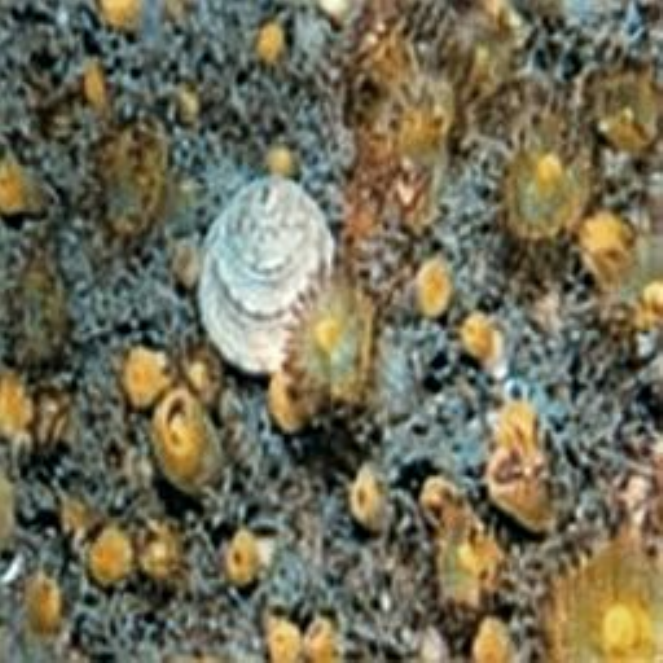}
	&\includegraphics[width=0.122\textwidth,height=0.075\textheight]{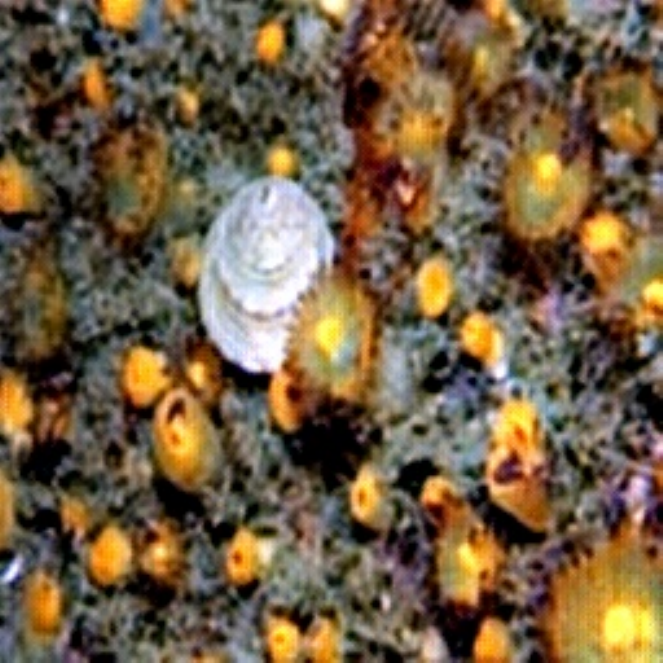}
	\\
	\includegraphics[width=0.122\textwidth]{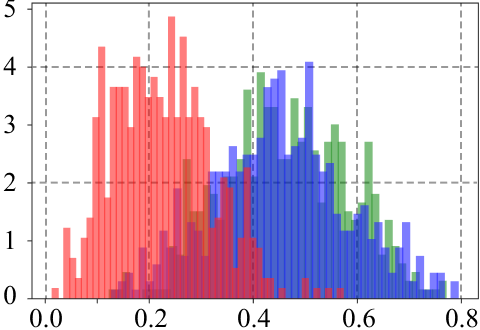}
	&\includegraphics[width=0.122\textwidth]{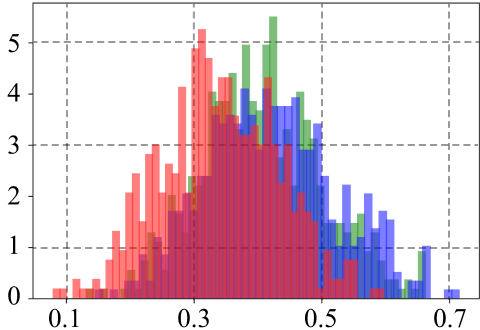}
	&\includegraphics[width=0.122\textwidth]{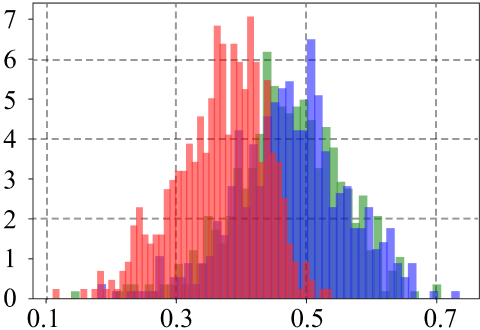}
	&\includegraphics[width=0.122\textwidth]{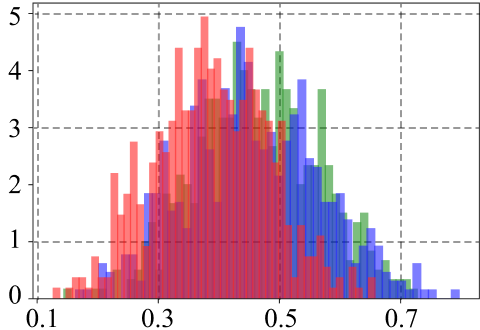}
	&\includegraphics[width=0.122\textwidth]{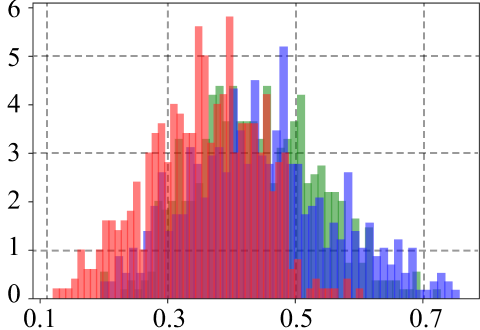}
	&\includegraphics[width=0.122\textwidth]{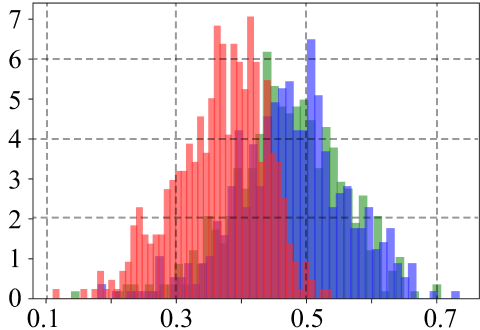}
	&\includegraphics[width=0.122\textwidth]{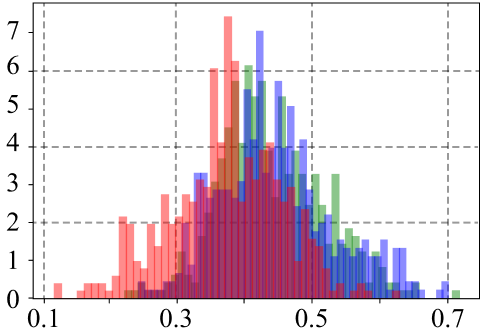}
	&\includegraphics[width=0.122\textwidth]{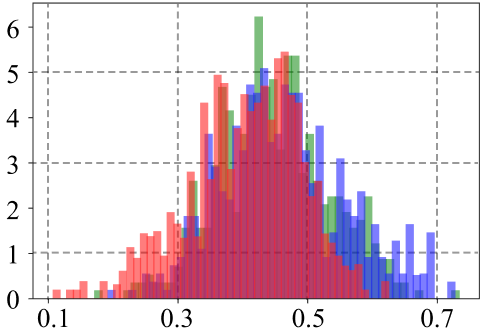}
\\ Input&Water-Net&DLIFM&Ucolor&TOPAL&TACL&SemiUIR&Ours
\end{tabular}
\vspace{-10pt}
\caption{Enhancement results on EUVP dataset. We calculate the histogram distribution of RGB color space in the dataset. The x and y axis of the diagram represents the pixel intensity and the probability distribution. It can be found that the distribution of RGB color space in our results fits to the distribution of in-air images best.}
\label{fig:EUVP}
\end{figure*}
\begin{figure*}[!htb]
	\centering
	\setlength{\tabcolsep}{1pt}
	\begin{tabular}{cccccccccccc}
		\includegraphics[width=0.122\textwidth,height=0.075\textheight]{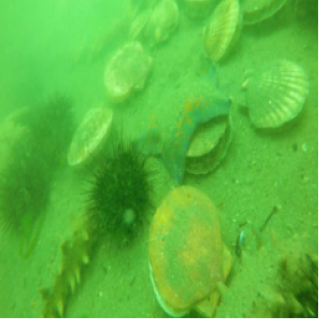}
		&\includegraphics[width=0.122\textwidth,height=0.075\textheight]{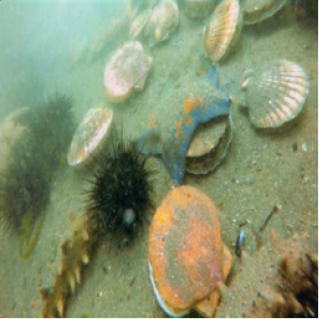}
		&\includegraphics[width=0.122\textwidth,height=0.075\textheight]{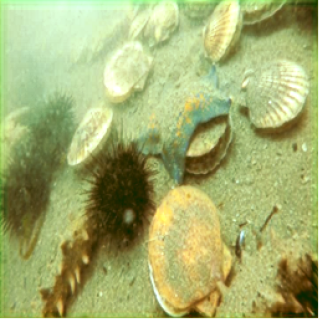}
		&\includegraphics[width=0.122\textwidth,height=0.075\textheight]{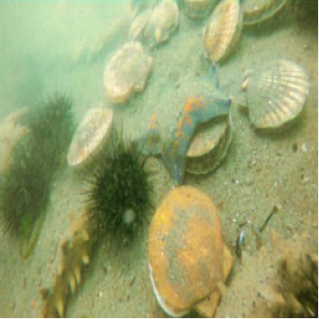}
		&\includegraphics[width=0.122\textwidth,height=0.075\textheight]{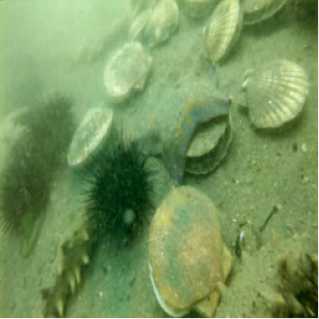}
		&\includegraphics[width=0.122\textwidth,height=0.075\textheight]{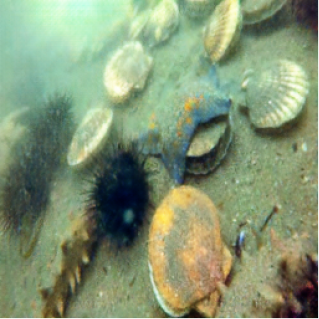}
		&\includegraphics[width=0.122\textwidth,height=0.075\textheight]{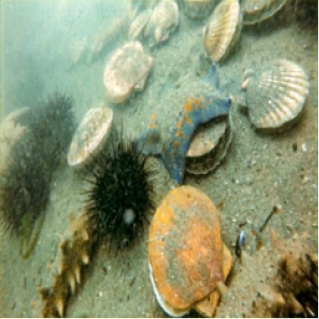}
		&\includegraphics[width=0.122\textwidth,height=0.075\textheight]{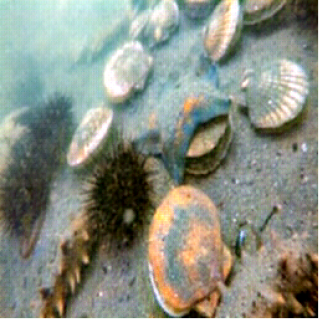}
		\\
		\includegraphics[width=0.122\textwidth,height=0.071\textheight]{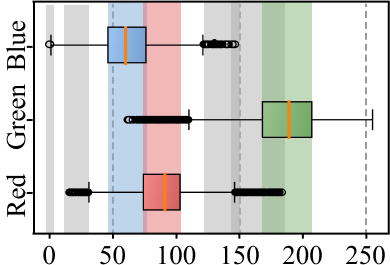}
		&\includegraphics[width=0.122\textwidth,height=0.071\textheight]{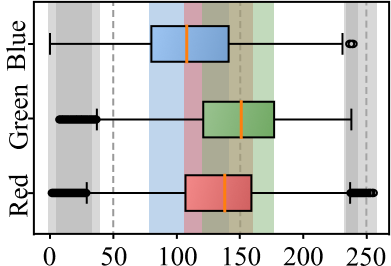}
		&\includegraphics[width=0.122\textwidth,height=0.071\textheight]{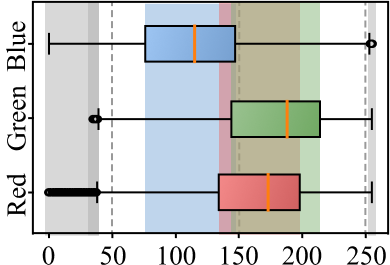}
		&\includegraphics[width=0.122\textwidth,height=0.071\textheight]{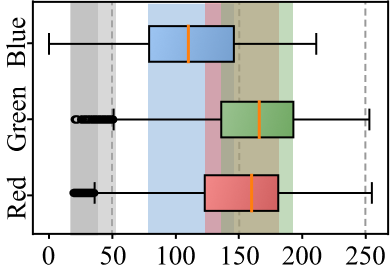}
		&\includegraphics[width=0.122\textwidth,height=0.071\textheight]{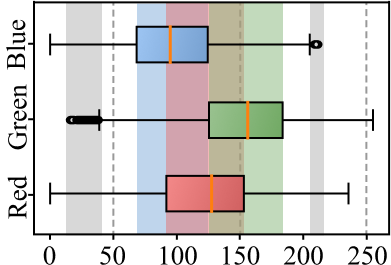}
		&\includegraphics[width=0.122\textwidth,height=0.071\textheight]{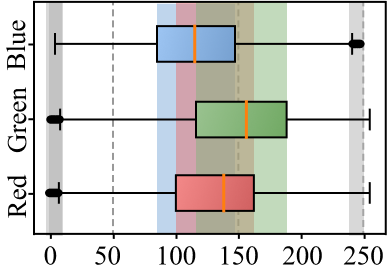}
		&\includegraphics[width=0.122\textwidth,height=0.071\textheight]{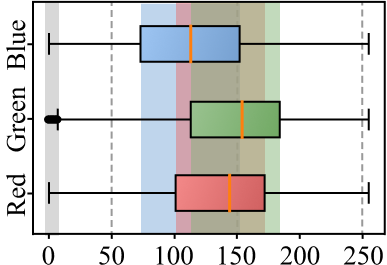}
		&\includegraphics[width=0.122\textwidth,height=0.071\textheight]{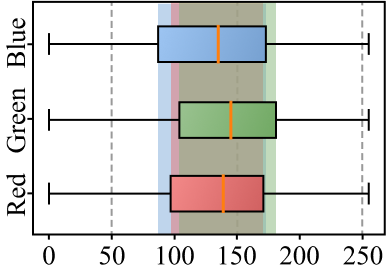}
		\\ Input&Water-Net&DLIFM&Ucolor&TOPAL&TACL&SemiUIR&Ours
	\end{tabular}
	\vspace{-10pt}
	\caption{Enhancement results on U45 dataset. We calculate the dispersion of RGB color space. The red, green and blue boxs respectively represent the corresponding color channel. It can be observed that the proposed results solve the problem of varying degrees of light decay underwater.}
\label{fig:U45}
\end{figure*}

\begin{figure*}[!htb]
\centering
\setlength{\tabcolsep}{1pt}
\begin{tabular}{cccccccccccc}
\includegraphics[width=0.122\textwidth,height=0.088\textheight]{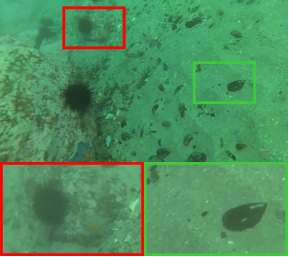}
&\includegraphics[width=0.122\textwidth,height=0.088\textheight]{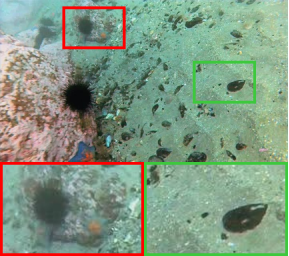}
&\includegraphics[width=0.122\textwidth,height=0.088\textheight]{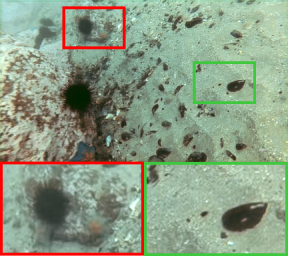}
&\includegraphics[width=0.122\textwidth,height=0.088\textheight]{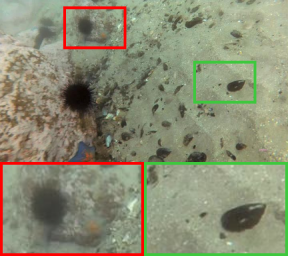}
&\includegraphics[width=0.122\textwidth,height=0.088\textheight]{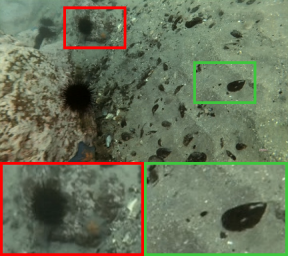}
&\includegraphics[width=0.122\textwidth,height=0.088\textheight]{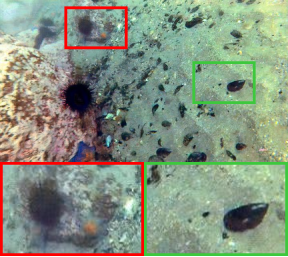}
&\includegraphics[width=0.122\textwidth,height=0.088\textheight]{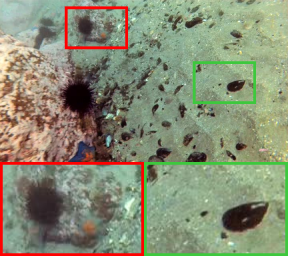}
&\includegraphics[width=0.122\textwidth,height=0.088\textheight]{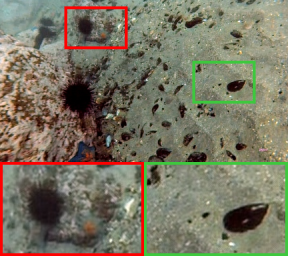}
\\
\includegraphics[width=0.122\textwidth,height=0.088\textheight]{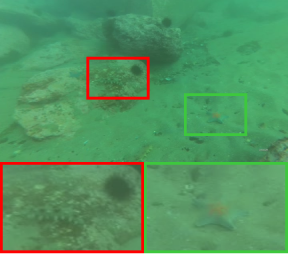}
&\includegraphics[width=0.122\textwidth,height=0.088\textheight]{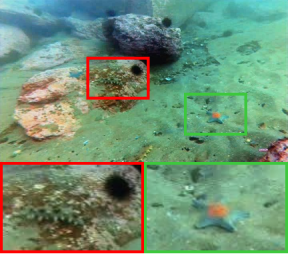}
&\includegraphics[width=0.122\textwidth,height=0.088\textheight]{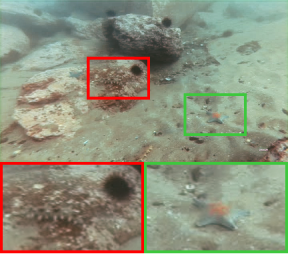}
&\includegraphics[width=0.122\textwidth,height=0.088\textheight]{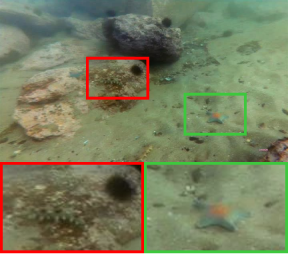}
&\includegraphics[width=0.122\textwidth,height=0.088\textheight]{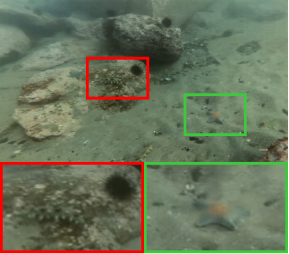}
&\includegraphics[width=0.122\textwidth,height=0.088\textheight]{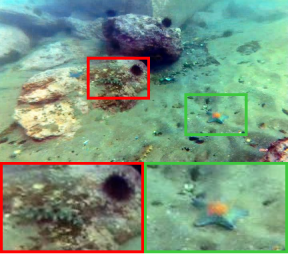}
&\includegraphics[width=0.122\textwidth,height=0.088\textheight]{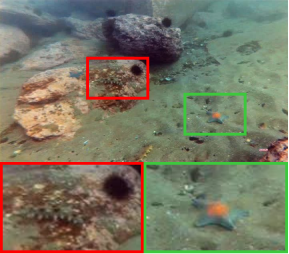}
&\includegraphics[width=0.122\textwidth,height=0.088\textheight]{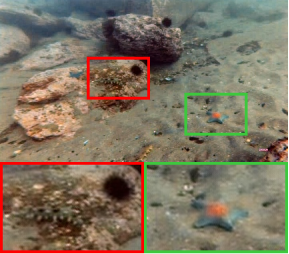}
\\Input&Water-Net&DLIFM&Ucolor&TOPAL&TACL&SemiUIR&Ours\\
\end{tabular}
\vspace{-10pt}
\caption{Enhancement results on UCCS dataset. Our method is obviously superior to other methods in correcting color attenuation and enhancing contrast, especially in the regions of red and green frames.}
\label{fig:UCCS_enhance}
\end{figure*}
\begin{figure}[!htb]
	\centering
	\setlength{\tabcolsep}{1pt}
	\begin{tabular}{cccc}
		\includegraphics[width=0.115\textwidth]{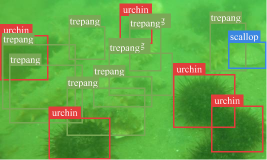}
		&\includegraphics[width=0.115\textwidth]{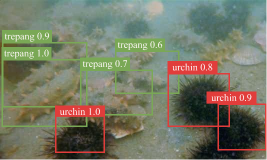}
		&\includegraphics[width=0.115\textwidth]{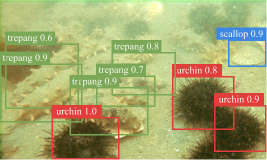}
		&\includegraphics[width=0.115\textwidth]{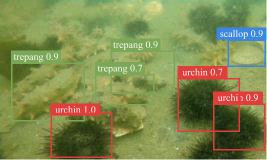}
		\\Ground Truth&Water-Net&DLIFM&Ucolor\\
		\includegraphics[width=0.115\textwidth]{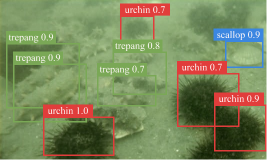}
		&\includegraphics[width=0.115\textwidth]{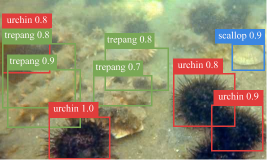}
		&\includegraphics[width=0.115\textwidth]{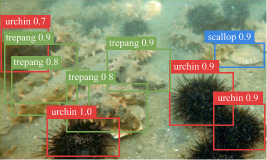}
		&\includegraphics[width=0.115\textwidth]{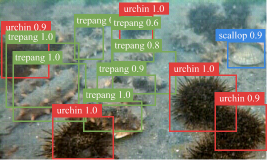}
		\\TOPAL&TACL&SemiUIR&Ours\\
	\end{tabular}
	\vspace{-11pt}
	\caption{ Evaluation of object detection on UCCS dataset. It can find that the proposed method achieves good performance in both accuracy and quantity.}
	\label{fig:UCCSdetection}
\end{figure}

\begin{figure*}[!htb]
\centering
\setlength{\tabcolsep}{0.95pt}
\begin{tabular}{cccccccc}
\includegraphics[width=0.2405\textwidth,height=0.133\textheight]{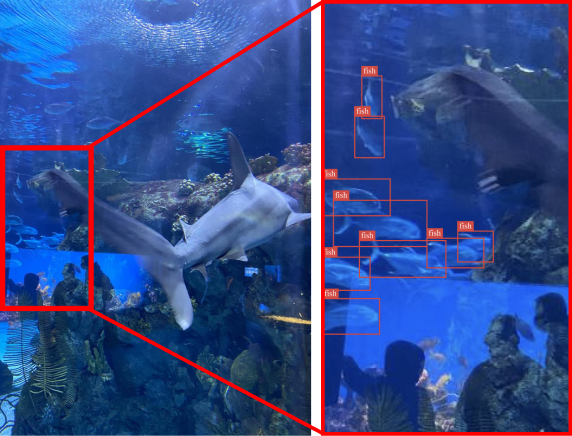}
&\includegraphics[width=0.1035\textwidth,height=0.133\textheight]{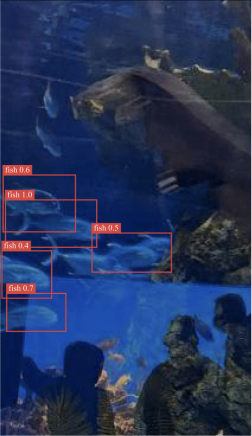}
&\includegraphics[width=0.1035\textwidth,height=0.133\textheight]{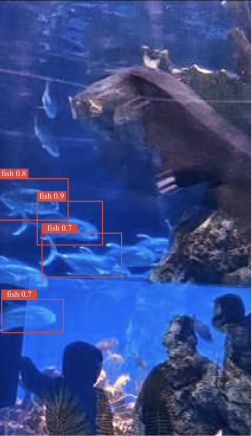}
&\includegraphics[width=0.1035\textwidth,height=0.133\textheight]{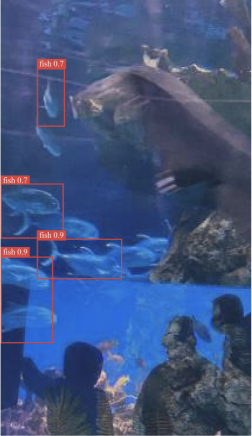}
&\includegraphics[width=0.1035\textwidth,height=0.133\textheight]{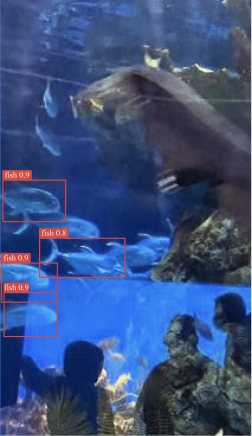}
&\includegraphics[width=0.1035\textwidth,height=0.133\textheight]{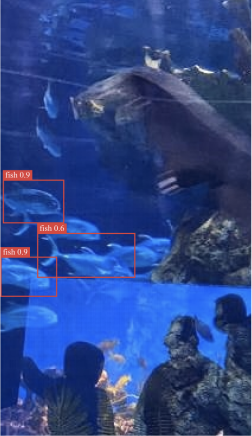}
&\includegraphics[width=0.1035\textwidth,height=0.133\textheight]{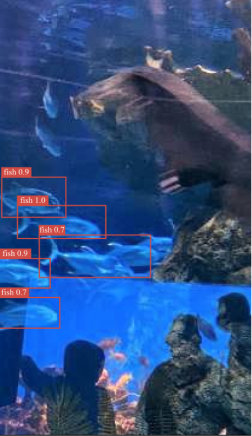}
&\includegraphics[width=0.1035\textwidth,height=0.133\textheight]{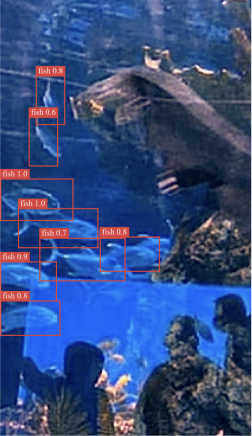}
\\Ground Truth&Water-Net&DLIFM&Ucolor&TOPAL&TACL&SemiUIR&Ours\\
\end{tabular}
\vspace{-12pt}
\caption{Evaluation of object detection on Aquarium dataset with different enhancement methods. It can find that the proposed method is most suitable for underwater object detection.}
\label{fig:Aquarium1}
\end{figure*}

\begin{table*}[h]
	\renewcommand\arraystretch{1.20}
	\caption{Quantitative comparison for underwater image enhancement in terms of UCIQE($\uparrow$), UIQM($\uparrow$) ,UICM($\uparrow$) ,PSNR($\uparrow$) and SSIM($\uparrow$). $\uparrow$ denotes that large values mean better results. The best and second results are marked in \textbf{bold} and \underline{underline}.}
	\vspace{-10pt}
	\setlength{\tabcolsep}{0.9mm}{
		\begin{tabular}{l|ccccc|ccc|ccc|ccc}
			\specialrule{1.1pt}{0pt}{0pt}
			\multirow{2}{*}{\textbf{Method}}
			&\multicolumn{5}{c|}{\cellcolor[RGB]{255,204,201}\textbf{UIEBD}}
			&\multicolumn{3}{c|}{\cellcolor[RGB]{154,255,154}\textbf{EUVP}}
			&\multicolumn{3}{c|}{\cellcolor[RGB]{93,173,226}\textbf{U45}}  
			&\multicolumn{3}{c}{\cellcolor[RGB]{225,255,154}\textbf{UCCS}}            \\
			& \textbf{UCIQE}&\textbf{UIQM}&\textbf{UICM}&\textbf{PSNR}&\textbf{SSIM}
			&  \textbf{UCIQE}&\textbf{UIQM}&\textbf{UICM}
			&  \textbf{UCIQE}&\textbf{UIQM}&\textbf{UICM}
			&  \textbf{UCIQE}&\textbf{UIQM}&\textbf{UICM}
			\\
			\hline 
			Water-Net
			& 0.5856            & 1.8681           & \underline{-37.9812}    
			& 18.8156           & 0.8257
			& 0.5836            & \textbf{4.4419}  & -17.2934
			& 0.5680            & 4.3612           & -23.7057
			& 0.5559            & 3.5007           & -14.0720
			\\DLIFM    
			& 0.6095            & 1.9553            & -38.0444   
			& 20.0894           & 0.8565
			& \underline{0.6205}& 4.1051           & -26.4611
			& 0.5880            & 4.2763           & -25.4130
			& 0.5200            & 2.9137           & -25.6534
			\\Ucolor   
			& 0.5542            & 1.6799      & -40.9628      
			& 19.7202           & 0.8273
			& 0.5846            & 4.3043           & -19.4697  
			& 0.5641            & 4.3708           & -21.6621 
			& 0.5201            & 3.4145           & -34.9304 
			\\TOPAL  
			& 0.5646            & 2.2013        & -48.1846       
			& 19.1631           & 0.8043         
			& 0.6010            & 4.2689           & -21.6535          
			& 0.5524            & 3.9625           & -32.6012
			& 0.4801            & 3.7760           & -23.3429
			\\TACL
			& 0.6128            &\underline{2.2848}&-39.8192         
			& 21.4864           & 0.8398 
			& 0.5842            & 3.6547           &-42.8193        
			& \underline{0.6223}& 4.4013           &-20.0128
			& \textbf{0.5935}   & 3.6249           &-18.3285
			\\SemiUIR    
			& \textbf{0.6202}   &1.9916            &-40.0554  
			& \underline{21.6818}  & \textbf{0.8753} 
			& 0.6175            &4.4360            &\underline{-15.3805}       
			& 0.6110            &\textbf{4.5201}   &\underline{-18.7781}
			& 0.5538            &\underline{3.7847}&\underline{-5.0835}
			\\Ours     
			&\underline{0.6157} &\textbf{2.5968}   &\textbf{-35.1569}
			&\textbf{21.7420} & \underline{0.8584} 
			&\textbf{0.6397}    &\underline{4.4400}&\textbf{-12.6861}  
			&\textbf{0.6229}    &\underline{4.4129}&\textbf{-13.6104}
			&\underline{0.5569} &\textbf{3.9842}&\textbf{-4.1614}
			\\ 
			\specialrule{1.1pt}{0pt}{0pt}
	\end{tabular}}
	\label{tab:evaluation_enhance}
\end{table*}

\begin{figure*}[!htb]
\centering
\setlength{\tabcolsep}{1.0pt}
\includegraphics[width=1.0\textwidth,height=0.157\textheight]{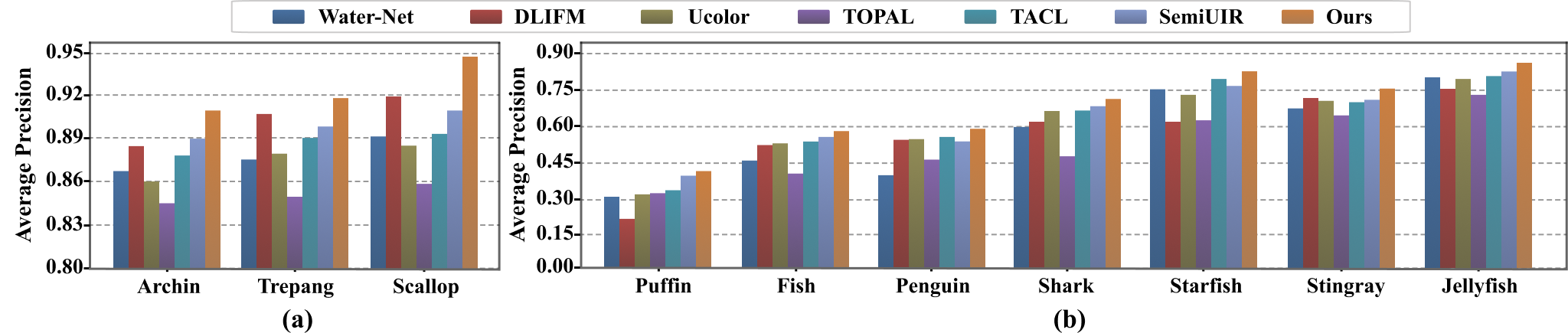}
\vspace{-22pt}
\caption{Detection accuracy on UCCS (a) and Aquarium (b) datasets with the enhanced images generating from the representative methods. The x-axis represents the category of the object.}
\label{fig:detetion_duibi2}
\end{figure*}
\begin{figure}[!htb]
\centering
\setlength{\tabcolsep}{1pt}
\begin{tabular}{cccccc}
\includegraphics[width=0.09\textwidth,height=0.075\textheight]{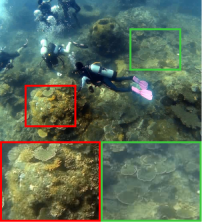}
&\includegraphics[width=0.09\textwidth,height=0.075\textheight]{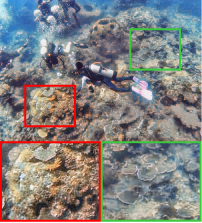}
&\includegraphics[width=0.09\textwidth,height=0.075\textheight]{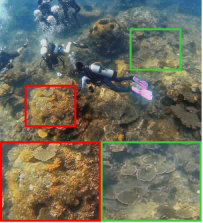}
&\includegraphics[width=0.09\textwidth,height=0.075\textheight]{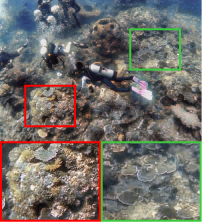}
&\includegraphics[width=0.09\textwidth,height=0.075\textheight]{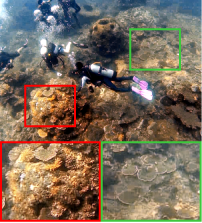}\\
\includegraphics[width=0.09\textwidth,height=0.075\textheight]{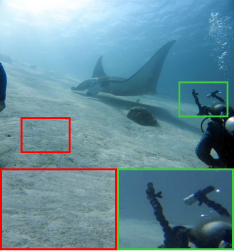}
&\includegraphics[width=0.09\textwidth,height=0.075\textheight]{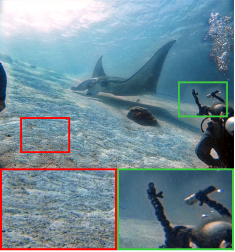}
&\includegraphics[width=0.09\textwidth,height=0.075\textheight]{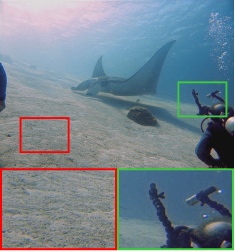}
&\includegraphics[width=0.09\textwidth,height=0.075\textheight]{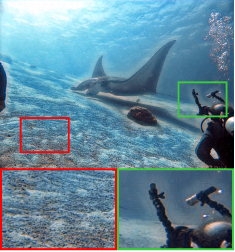}
&\includegraphics[width=0.09\textwidth,height=0.075\textheight]{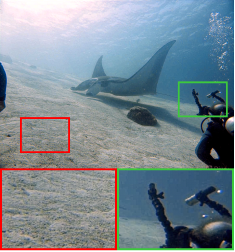}
\\Input&w/o $\mathcal{L}_1$&w/o  $\mathcal{L}_s$& w/o $\mathcal{L}_c$&Ours\\
\end{tabular}
\vspace{-11pt}
\caption{Ablation study of loss function.~$\mathcal{L}_1$, $\mathcal{L}_s$ and  $\mathcal{L}_c$ denote L1 loss, style loss, and contrasive loss.}
\label{fig:visualizationforlossablation}
\end{figure}

\begin{figure*}[!htb]
	\centering
	\setlength{\tabcolsep}{0.10pt}
	\includegraphics[width=\textwidth,height=0.11\textheight]{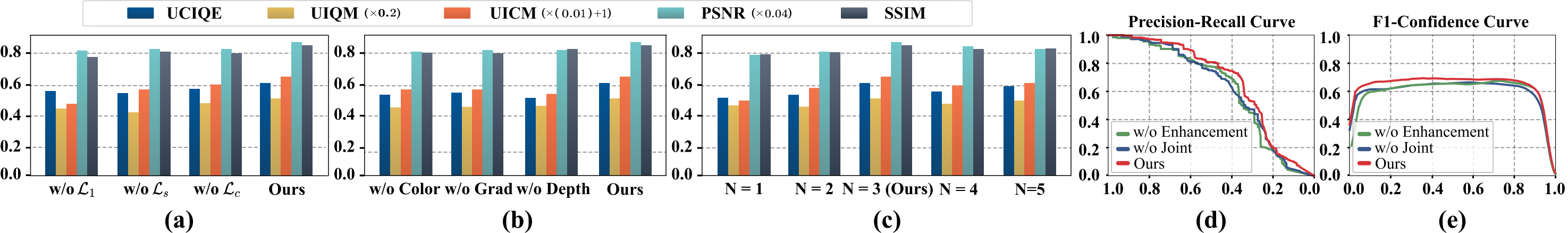}
	\vspace{-20pt}
	\caption{Quantitative results of the ablation experiment. (a) denotes the evaluation on the loss function. (b) represents the evaluation for the input of Heuristic Prior guided Encoder. (c) represents the evaluation for the number of Hybrid Invertible Blocks. (d) denotes the Precision-Recall curve of ablation study for underwater object detection, the x and y axis respectively represent recall and precision. (e) denotes the F1-confidence curve, the x and y axis represent the confidence and F1 score.}
	\label{fig:lossablation}
\end{figure*}

\begin{figure}[!htb]
\centering
\setlength{\tabcolsep}{1pt}
\begin{tabular}{ccccc}
\includegraphics[width=0.09\textwidth,height=0.077\textheight]{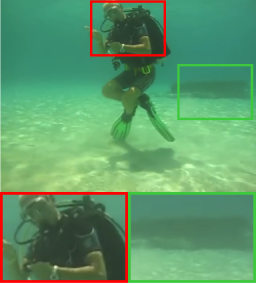}
&\includegraphics[width=0.09\textwidth,height=0.077\textheight]{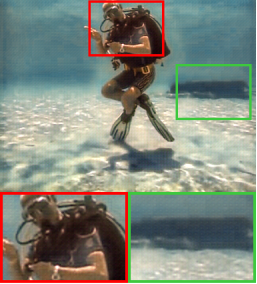}
&\includegraphics[width=0.09\textwidth,height=0.077\textheight]{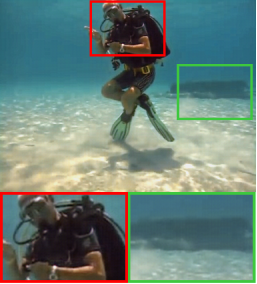}
&\includegraphics[width=0.09\textwidth,height=0.077\textheight]{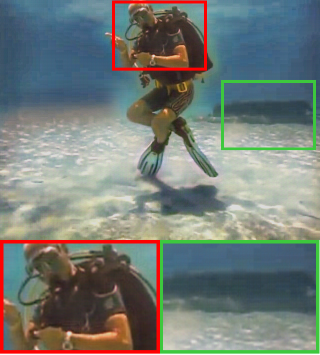}
&\includegraphics[width=0.09\textwidth,height=0.077\textheight]{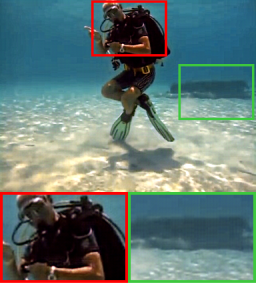}\\
\includegraphics[width=0.09\textwidth,height=0.077\textheight]{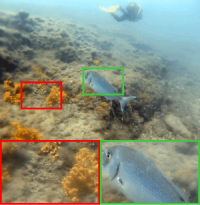}
&\includegraphics[width=0.09\textwidth,height=0.077\textheight]{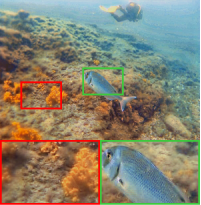}
&\includegraphics[width=0.09\textwidth,height=0.077\textheight]{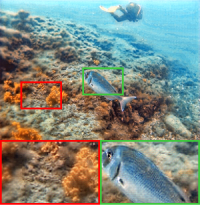}
&\includegraphics[width=0.09\textwidth,height=0.077\textheight]{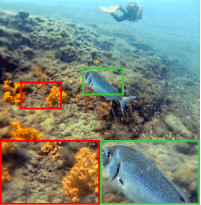}
&\includegraphics[width=0.09\textwidth,height=0.077\textheight]{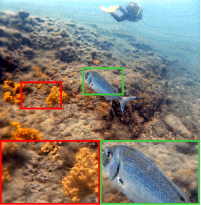}
\\Input&w/o Color&w/o Grad&w/o Depth&Ours\\
\end{tabular}
\vspace{-12pt}
\caption{Visualization for ablation study on the input of Heuristic Prior guided Encoder.}
\label{fig:visualizationforinputencoder}
\end{figure}

\begin{figure}[!htb]
	\centering
	\setlength{\tabcolsep}{1pt}
	\begin{tabular}{ccc}
		\includegraphics[width=0.151\textwidth,height=0.058\textheight]{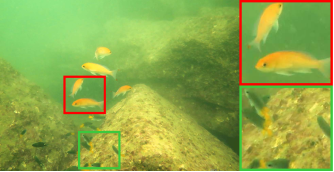}
		&\includegraphics[width=0.151\textwidth,height=0.058\textheight]{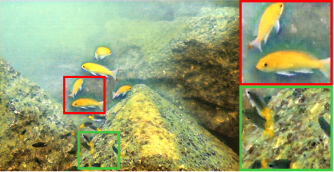}
		&\includegraphics[width=0.151\textwidth,height=0.058\textheight]{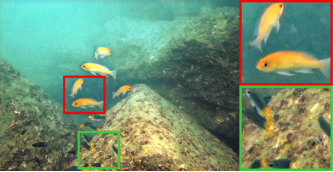}
		\\Input&$N$ = 1&$N$ = 2\\
		\includegraphics[width=0.151\textwidth,height=0.058\textheight]{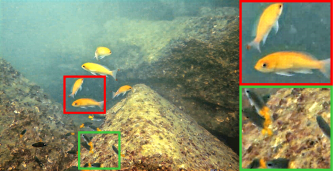}
		&\includegraphics[width=0.151\textwidth,height=0.058\textheight]{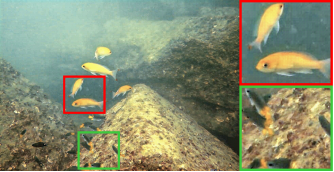}
		&\includegraphics[width=0.151\textwidth,height=0.058\textheight]{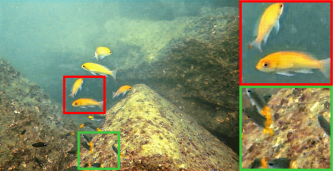}
		\\$N$ = 3& $N$ = 4& $N$ = 5\\
	\end{tabular}
	\vspace{-12pt}
	\caption{Visualization for ablation study on the number of Hybrid Invertible Blocks.}
	\label{fig:visualizationforblocks}
\end{figure}

\section{Experiments}
In this section, we evaluate the effectiveness of the proposed method through qualitative and quantitative comparison. Specifically, widely used UIEBD~\cite{li2019waternetuiebd}, EUVP~\cite{islam2020euvp}, U45~\cite{li2019fusion} and UCCS~\cite{liu2020uccs} datasets are used to evaluate the performance of WaterFlow for underwater image enhancement. Six representative methods Water-Net~\cite{li2019waternetuiebd}, DLIFM~\cite{chen2021dlifm}, Ucolor~\cite{li2021ucolor}, TOPAL~\cite{jiang2022topal}, TACL~\cite{liu2022tacl} and SemiUIR~\cite{huang2023contrastive} in recent years are compared for evaluating the performance. Both subjective and objective results are adopted for analysis. Underwater Color Image Quality Evaluation (UCIQE)~\cite{yang2015uciqe}, Underwater Image Quality Measurement (UIQM)~\cite{panetta2015uiqmuism} and Underwater Image Contrast Measure (UICM)~\cite{panetta2015uiqmuism} are used as non-reference evaluation metrics. A higher UCIQE, UIQM or UICM score indicates a better human visual perception. Peak Signal to Noise Ratio (PSNR) and Structural Similarity (SSIM)~\cite{wang2004image} are adopted as the fully reference evaluations. Higher values of these metrics indicate a better similarity between the resulting image and the reference image in terms of both content and structure. 
Furthermore, UCCS~\cite{liu2020uccs} and Aquarium~\cite{aquarium} datasets are used to evaluate the ability of the proposed method to adapt to subsequent detection tasks. The commonly utilized Average Precision (AP) is adopted as the detection-driven evaluation metric, which is positive for the performance of underwater object detection. 
\subsection{Implement Details}
Our network is implemented by the pytorch and trained on NVIDIA RTX 3090 GPU. We first randomly crop the UIEBD~\cite{li2019waternetuiebd} dataset into the size of $384 \times 384$ and use the cropped dataset as the training dataset for underwater image enhancement. UCCS~\cite{liu2020uccs} and Aquarium~\cite{aquarium} datasets are used as the training  dataset for underwater object detection. We first train the heuristic normalizing flow and Detection Perception Module separately for $5\times e^{5}$ iterations. Then, we jointly train the pre-trained modules for $3 \times e^{5}$ iterations. During the training process, Adam was used as our optimizer with the uniform learning rate of 1e-6 and the batch size of 2. $\lambda_1$, $\lambda_2$, $\lambda_3$, $\lambda_4$ are set to $1$, $100$, $0.1$, $1$.
\subsection{Qualitative Results}
For visual comparison, Fig.~\ref{fig:UIEBD} shows the enhancement results of all methods on UIEBD dataset. Water-Net and Ucolor not only fail to recover the original reflection, but also suffer from low contrast and saturation. DLIFM, TOPAL, TACL and SemiUIR have limited effects on underwater image enhancement, which still remain conspicuous color deviation at close scenes. Compared with the other methods, the proposed method recovers the degradation of light and rectifies the unexpected coloration. In addition, we demonstrate the pixel distribution similarity diagram of a particular region between the corresponding enhanced and reference images. The red line and the blue line in the diagram indicate the pixel distribution of the reference image and the enhanced image respectively. Notably, the proposed method yields enhanced results closest to the reference image.

The comparison on the EUVP dataset is shown in Fig.~\ref{fig:EUVP}. Ucolor and TOPAL introduce artifacts and unnatural colors. Water-Net, TACL and SemiUIR fail to alleviate the scattering of light underwater. Although DLIFM produces pleasing images, the instructive details of the image have not been recovered well. Compared with the other methods, the proposed method not only effectively corrects the color deviation, but also restores the scene radiance and contrast of the image. Fig.~\ref{fig:EUVP} also shows the distribution of RGB color space for the images obtained by different methods. In general, the proposed method more effectively addresses the swift attenuation of red wavelengths underwater, with its color space distribution closely approximating that of in-air images.

The qualitative comparison on the U45 datasets is shown in Fig.~\ref{fig:U45}. It is obvious that the red light and blue light decay significantly more than the green light underwater in this acquisition environment. However, the other compared methods exhibit limited effectiveness in restoring the decay of blue light. The boxplot for the intensity distribution of the RGB color channels is set below each image, with the horizontal axis representing pixel intensity. Additionally, semi-transparent rectangles are introduced to expound the relationship between distinct color spaces more lucidly, with the gray rectangle signifying the region where outliers emerged. In comparison with the proposed method, the other methods not only have a noticeably inferior ability to revive the attenuation of blue light but also introduce certain aberrant outliers, adversely impacting the efficacy of the enhanced images.

The visual comparison on the UCCS datasets is shown in Fig.~\ref{fig:UCCS_enhance}. TOPAL fails to restore the color deviation of the image, which suffers from low saturation. Unreasonable color artifacts are introduced by Water-Net, TACL and SemiUIR. Informative details are ignored by DLIFM and Ucolor, resulting in obscure reflections and low contrast. Compared with other methods, our method significantly restores the brightness and contrast of underwater scenes without the interference introduction.

We utilize the same comparison methods to assess whether the proposed enhancement results implicitly contain more semantic features that are applicable to subsequent detection tasks. The visual comparison of UCCS dataset is shown in Fig.~\ref{fig:UCCSdetection}. The proposed method achieves a higher detection confidence than other methods with more detected trepangs and urchins. The qualitative comparison of the Aquarium dataset is shown in Fig.~\ref{fig:Aquarium1}, which clearly shows the remarkable advantages of the proposed method in adapting to subsequent detection tasks. 
\subsection{Quantitative Results}
The quantitative results of representative methods for underwater image enhancement are shown in Tab.~\ref{tab:evaluation_enhance}. It can be observed that the proposed method exceeds all other methods in UICM, and achieves competitive results in UCIQE and UIQM. However, as noted by~\cite{berman2021,huang2023contrastive}, although both UCIQE and UIQM metrics can represent the degree of restoration of underwater images, they are somewhat heuristic and have limited applicability on evaluating the performance of underwater image enhancement. We further compared fully-reference evaluation on the UIEBD dataset which contains reference images. 
The proposed method achieves a significant advantage on PSNR and is only lower than SemiUIR on SSIM. However, as noted by~\cite{li2021ucolor}, the reference images of UIEBD are synthetic by multiple enhancement methods, which indicates that some unreliable samples may affect the evaluation of the performance. Combining with NR-IQA and full-reference benchmarks, our method still achieves great advantages in underwater image enhancement, consistent with the qualitative results.
The quantitative comparison for underwater object detection is shown in  Fig.~\ref{fig:detetion_duibi2}. It can be observed that the proposed method outperforms the other competitive methods in terms of detection accuracy in both UCCS and Aquarium datasets. It can be proven that the proposed results can not only achieve good visual enhancement effects, but also potentially contain more perceptual information which is beneficial for subsequent detection tasks.

\subsection{Ablation Study}
\subsubsection{Study on Loss Function}
We discuss the effectiveness of different loss functions. The visual results are shown in Fig.~\ref{fig:visualizationforlossablation}. w/o  $\mathcal{L}_1$ and w/o  $\mathcal{L}_c$ make the images deviate from their intrinsic colors and introduce unpleasant artifacts.  
w/o $\mathcal{L}_s$ still suffers from low contrast and clarity although it has improved the color balance of the image to a certain extent. Combined with quantitative comparison in (a) of Fig.~\ref{fig:lossablation}, the proposed method achieves the best enhancement effect, which verifies the effectiveness of all functions.
\subsubsection{Study on Heuristic Prior guided Encoder}
We conduct ablation experiments on the input of HPE to verify the effectiveness of each input to the proposed method. The qualitative comparison is shown in Fig.~\ref{fig:visualizationforinputencoder}. In the first example, when the color image or depth map is not adopted as input, unpleasant shadows are introduced. The effect of image enhancement is obviously limited when the gradient map is not used. In the second example, the proposed method achieved the best enhancement effect, which not only restore the color of the ground  but also significantly improved the clarity of the content in the red and green frames. The quantitative comparison is shown in (b) of Fig.~\ref{fig:lossablation}, which shows that the proposed method achieves the best results on all benchmarks. Any input removed significantly attenuates the enhanced image.
\subsubsection{Study on Number of Hybrid Invertible Blocks}
We conduct ablation experiments on the number of Hybrid Invertible Blocks $N$. The visual comparison is illustrated in Fig.~\ref{fig:visualizationforblocks}. It is obvious that when the number of blocks is less than 3, the enhanced image suffers from local over enhancement and the image artifacts introduction. Quantitative comparison is shown in (c) of Fig.~\ref{fig:lossablation}. Considering the limitations of network parameters and computational costs, we finally employ $N$ as 3 in our framework.
\subsubsection{Study on underwater object detection}
In order to verify the effectiveness of the proposed method for underwater object detection. We directly train the existing detection network without enhancement on underwater images to verify the effect of enhanced data on underwater object detection. At the same time, the enhancement module and the Detection Perception Module are not trained jointly, so as to test whether DPM can make the enhanced data more conducive to subsequent detection tasks. Quantitative comparison is shown in (d) and (e) of Fig. \ref{fig:lossablation}. w/o Enhancement and w/o Joint respectively indicate that the enhanced image is not used as a preprocessing stage and the latent features are not transmitted to the enhancement module from the DPM. Evidently, the proposed method achieves the highest detection performance, which proves the effectiveness of the DPM.

\section{CONCLUSION}
In this paper, we propose a detection-driven heuristic normalizing flow underwater image enhancement. We first adopt the underwater image restoration process into the normalizing flow to establish the invertible mapping of the degraded image and the its clear counterpart through bilateral constraints. Then we further introduce the heuristic prior guided injector to improve the representation ability of the enhancement network by progressively estimating the underwater imaging parameters. At last, we introduce the detection perception module, which propagates high-level perceptual features into the enhancement model as gradient information to help generate the detection-driven enhanced image. Extensive experiments on multiple benchmarks show that the proposed method not only achieves good visual enhancement effects, but is also more suitable for subsequent detection tasks.

\section*{ACKNOWLEDGMENTS}
This work is partially supported by the National Key R\&D Program of China (No. 2022YFA1004101), the National Natural Science Foundation of China (No. U22B2052).
\newpage
\balance
\bibliographystyle{ACM-Reference-Format}
\bibliography{sample-base}

\end{document}